\documentclass{article} 
\usepackage{iclr2025_conference,times}

\usepackage{amsmath,amsfonts,bm}

\def\eqref#1{equation~\ref{#1}}

\def\1{\bm{1}}

\DeclareMathAlphabet{\mathsfit}{\encodingdefault}{\sfdefault}{m}{sl}
\SetMathAlphabet{\mathsfit}{bold}{\encodingdefault}{\sfdefault}{bx}{n}

\usepackage{hyperref}
\usepackage{url}
\usepackage{booktabs}       
\usepackage{wrapfig}

\usepackage{tikz}

\usepackage[utf8]{inputenc} 
\usepackage[T1]{fontenc}    
\usepackage{amsfonts}       
\usepackage{nicefrac}       
\usepackage{microtype}      
\usepackage{xcolor}         
\usepackage{amsmath}
\usepackage{bm}
\usepackage{etoolbox, siunitx}        
\usepackage[normalem]{ulem}
\usepackage{graphicx}
\usepackage{amssymb}
\usepackage{xfrac}

\definecolor{mygreen}{rgb}{0.0,0.55,0.3}
\newcommand{\eg}{\emph{e.g.}}
\newcommand{\ie}{\emph{i.e.}}
\newcommand{\etal}{\emph{et al.}}

\usepackage{comment}


\title{Rethinking the role of frames for SE(3)-invariant crystal structure modeling}
\title{CrystalFramer: Rethinking the Role of Frames for SE(3)-Invariant Crystal Structure Modeling}

\author{%
Yusei Ito${}^{*1,2}$, %
Tatsunori Taniai${}^{*1}$, %
Ryo Igarashi${}^1$, %
Yoshitaka Ushiku${}^1$, %
and Kanta Ono${}^2$\\
${}^*$Contributed equally.  %
${}^1$OMRON SINIC X Corporation %
${}^2$Osaka University %
\\
\url{https://omron-sinicx.github.io/crystalframer/}
}

\author{
Yusei Ito${}^{*1,2}$
\And
Tatsunori Taniai${}^{*1}$
\And
Ryo Igarashi${}^1$
\And
Yoshitaka Ushiku${}^1$
\And
Kanta Ono${}^2$\\
\AND
\vspace{-28pt}\\
${}^*$Contributed equally.~~~~
${}^1$OMRON SINIC X Corporation~~~~
${}^2$Osaka University\\
\url{https://omron-sinicx.github.io/crystalframer/}
}

\iclrfinalcopy 
\begin{document}

\maketitle

\begin{abstract}
Crystal structure modeling with graph neural networks is essential for various applications in materials informatics, and capturing SE(3)-invariant geometric features is a fundamental requirement for these networks. A straightforward approach is to model with orientation-standardized structures through structure-aligned coordinate systems, or ``frames.'' However, unlike molecules, determining frames for crystal structures is challenging due to their infinite and highly symmetric nature. In particular, existing methods rely on a statically fixed frame for each structure, determined solely by its structural information, regardless of the task under consideration. Here, we rethink the role of frames, \emph{questioning whether such simplistic alignment with the structure is sufficient}, and propose the concept of \emph{dynamic frames}. While accommodating the infinite and symmetric nature of crystals, these frames provide each atom with a dynamic view of its local environment, focusing on actively interacting atoms. We demonstrate this concept by utilizing the attention mechanism in a recent transformer-based crystal encoder, resulting in a new architecture called CrystalFramer. Extensive experiments show that CrystalFramer outperforms conventional frames and existing crystal encoders in various crystal property prediction tasks.
\end{abstract}

\section{Introduction}
Geometric graph neural networks~\citep{xie18cgcnn,chen2019megnet,choudhary21alignn,lin2023potnet}, including transformer variants~\citep{yan22matformer,yan2024comformer,taniai2024crystalformer}, play a central role in machine learning (ML)-based structural modeling of materials. This technology offers a powerful alternative to conventional simulation methods, such as density functional theory (DFT) calculations, enabling high-throughput prediction of material properties. Furthermore, it serves as the basis for various ML applications in materials science, such as material embedding learning~\citep{Suzuki2022dml,suzuki2025contrastive,chiba2023nesf} and crystal generation~\citep{jiao2023crystalpredict}.

A key requirement for these networks is the ability to capture essential features of materials embedded in their crystal structures. Crystal structures are periodic, infinitely repeating arrangements of atoms in 3D space, typically represented by minimum repeatable patterns called unit cells. 
Material properties, such as formation energy and bandgap, are invariant under rigid transformations (\ie, rotations and translations) in crystal structures, as well as under variations in their unit cells. This fact leads to the so-called periodic SE(3) invariance~\citep{yan22matformer} as an essential property for crystal encoders.
Therefore, recent studies have explored various forms of richer yet invariant structural information beyond the simplest interatomic distances~\citep{chen22m3gnet,duval23faenet,yan2024comformer}. 

\begin{figure}[t]
\centering
\includegraphics[width=1.0\textwidth]{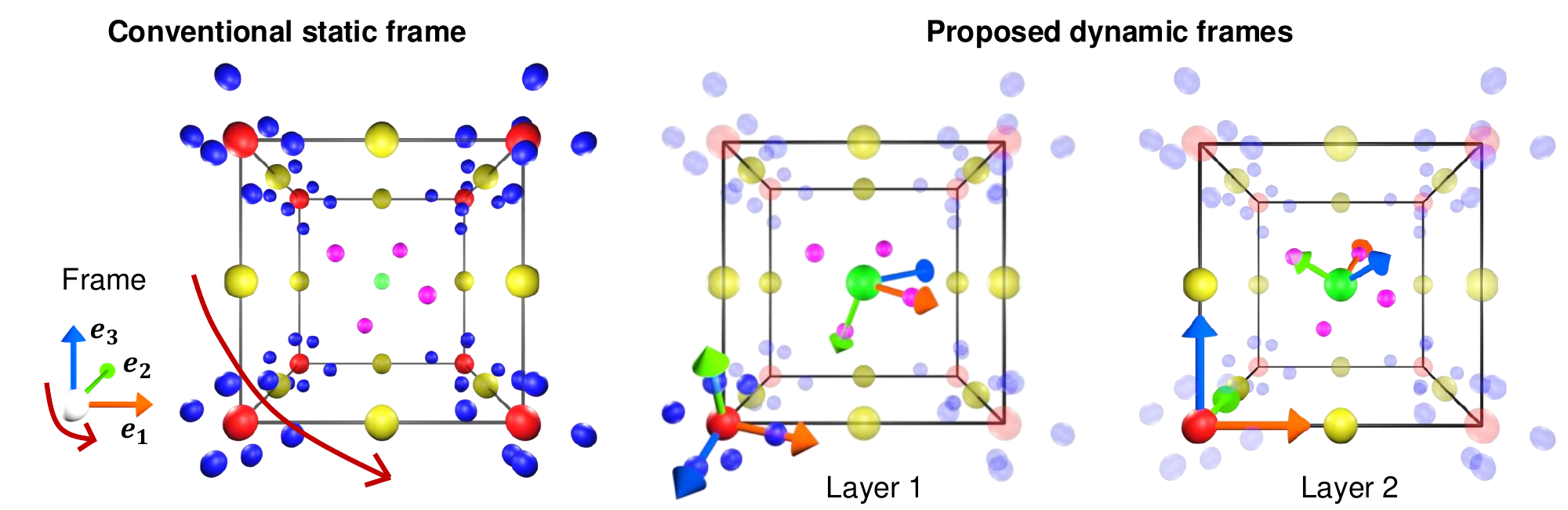}
\caption{\textbf{Conventional static frame and proposed dynamic frames.} Conventional frames are determined statically to align with the structure, ensuring consistency under rotation and providing a canonical global representation of the structure. This consistency is schematically illustrated by the curved arrows. By contrast, the proposed dynamic frames are determined for each atom in each message-passing layer, by considering the local dynamic environment around that atom in that layer.}
\label{fig:concept}
\end{figure}

One approach that has shown promising results for molecules~\citep{puny2022frameaveraging} is the use of ``frames.'' A frame is a coordinate system aligned equivariantly to a given structure to provide an orientation-standardized view of the structure (see Fig.~\ref{fig:concept}, left). 
Frames allow arbitrary networks to directly exploit rich 3D structural features, including the relative positions between atoms and their directions, without imposing any architectural constraints. 
However, determining frames for crystals is more challenging than for molecules, primarily due to the infinite and symmetric nature of crystals. 

In this work, we study a new family of frames for crystal structures in rethinking the role of frames. 
We hypothesize that \emph{the essential role of frames is not merely to provide a structure-aligned coordinate system for a given structure, but rather to align the coordinate system with the interatomic interactions acting on the structure}.
Following this belief, we propose a novel concept of \emph{dynamic frames}. These frames define local coordinate systems centered on individual atoms by dynamically accounting for the atoms actively engaged in learned interactions in each interatomic message-passing layer (Fig.~\ref{fig:concept}, right).
This concept challenges the conventional notion of `static frames,' which are based on the premise of providing fixed views of structures~\citep{puny2022frameaveraging}. Thus, whether such a dynamic frame is effective or not is an unexplored non-trivial question, which we aim to answer. %

To verify this concept, we develop several types of dynamic frames by utilizing the self-attention mechanism~\citep{taniai2024crystalformer} to quantify the interaction engagement. We conduct extensive comparisons on datasets derived from the JARVIS, Materials Project (MP), and Open Quantum Materials Database (OQMD). Our results show that our method outperforms existing frame methods for crystals~\citep{duval23faenet,yan2024comformer} and other state-of-the-art networks~\citep{choudhary21alignn,chen22m3gnet,yan22matformer,yan2024comformer,lin2023potnet,taniai2024crystalformer} across various crystal property prediction tasks. We release our code online.

\section{Preliminaries}
\subsection{Crystal structure}
\label{sec:crystal_structure}
A crystal structure is described by its 3D unit cell slice, denoted as $(A, P, L)$ following \cite{yan22matformer}. A unit cell is a parallelepipedal structure containing a finite number, say $N$, of atoms. The species (atomic numbers) and 3D Cartesian coordinates of these atoms are provided as $A = [a_1, a_2, ..., a_N] \in \mathbb{N}^{1 \times N}$ and $P = [\bm{p}_1, \bm{p}_2, ..., \bm{p}_N] \in \mathbb{R}^{3 \times N}$. The parallelepipedal cell shape is given by three vectors: $L=[\bm{l}_1, \bm{l}_2, \bm{l}_3] \in \mathbb{R}^{3 \times 3}$, called lattice vectors. By tiling the parallelepiped unit cell to fill 3D space, the species and positions of all the atoms in the crystal structure are determined as
\begin{align}
\hat{A}&=\{a_{i(\bm{n})} | a_{i(\bm{n})}=a_i, \bm{n}\in\mathbb{Z}^3, 1\leq i \leq N\},\\
\hat{P}&=\{\bm{p}_{i(\bm{n})} | \bm{p}_{i(\bm{n})}=\bm{p}_i+L\bm{n}, \bm{n}\in\mathbb{Z}^3, 1\leq i \leq N\}.
\end{align}
Following \cite{taniai2024crystalformer}, we use $i$ to denote the $i$-th atom in a unit cell, and  $i(\bm{n})$ to denote its duplicate by a unit-cell translation: $L\bm{n} = n_1\bm{\ell}_1 + n_2\bm{\ell}_2 + n_3\bm{\ell}_3$. We use $j$ and $j(\bm{n})$  similarly.

\subsection{Transformers for crystal structures}
\label{sec:crystalformer}
Geometric graph neural networks are used as crystal encoders in various materials-related tasks. These encoders typically represent the state of a given crystal structure by a set of atom-wise abstract state features, $X = [\bm{x}_1, \bm{x}_2, ..., \bm{x}_N] \in \mathbb{R}^{d \times N}$. These states are initially provided as atom embeddings,  $X^{(0)} \gets \text{AtomEmbedding}(A)$, which only symbolically represent atomic species. The encoders then evolve these states through interatomic message-passing layers, $X^{(t+1)} \gets f^t(X^{(t)}, P, L)$,
to eventually reflect the atomic states in the given structure appropriate for a target task.
Since the seminal work of \cite{xie18cgcnn} and \cite{schutt18schnet}, 
graph neural networks (GNNs) have long been the standard for crystal encoders until the advent of transformer-based networks by recent work~\citep{yan22matformer,yan2024comformer,taniai2024crystalformer}.

In particular, \cite{taniai2024crystalformer} have developed simple physics-informed formalism for crystal encoders using a self-attention mechanism. 
By imitating interatomic potential summations for energy calculations in physics simulations, they model the evolution of current state $\bm{x}$ using \emph{infinitely connected distance-decay attention}. This attention mechanism models the interactions between each unit-cell atom $i$ and all the infinitely repeating atoms $j(\bm{n})$ in the entire crystal structure as
\begin{equation}
\bm{x}'_i = \frac{1}{Z_i} {\sum_{j=1}^N\sum_{\bm{n}\in \mathbb{Z}^3} \exp\left(\frac{{\bm{q}_i^T \bm{k}_{j}}}{\sqrt{d_K}} - \frac{\|\bm{p}_{j(\bm{n})}-\bm{p}_i\|^2}{2\sigma_i^2}\right) \left(\bm{v}_{j} +\bm{\psi}_{ij(\bm{n})}\right)}. \label{eq:attention} 
\end{equation}
Here, query $\bm{q}$, key $\bm{k}$, and value $\bm{v}$ are linear projections of current state $\bm{x}$. Scalar $\sigma_i$ is a tail-length variable for Gaussian distance-decay attention, adaptively derived from $\bm{x}_i$. Vector $\bm{\psi}_{ij(\bm{n})}$ is a geometric position embedding that encodes the distance, $\|\bm{p}_{j(\bm{n})} - \bm{p}_i \|$, between atoms $i$ and $j(\bm{n})$.
Scalar $Z_i = \sum_j \sum_{\bm{n}} \exp({\bm{q}_i^T \bm{k}_{j}}/\sqrt{d_K} - \|\bm{p}_{j(\bm{n})}-\bm{p}_i\|^2/2\sigma_i^2)$ is the normalizer of softmax attention weights. The exponential distance-decay factor in Eq.~\ref{eq:attention} provably ensures its rapid convergence within a finite range of unit-cell shifts $\bm{n}$~\citep{taniai2024crystalformer}.

Their method, called Crystalformer, enjoys a good balance between a strong physically-motivated inductive bias and the flexibility of abstract feature representations. It is considered the state of the art with other GNN-based~\citep{lin2023potnet} and transformer-based~\citep{yan2024comformer} methods. 

We utilize Crystalformer as a baseline in this work. This is because its architecture closely follows the standard softmax attention~\citep{vaswani17transformer} and is suitable to demonstrate our concept of dynamic frames, while other existing transformers~\citep{yan22matformer,yan2024comformer} use distinct channel-wise sigmoid attention. We discuss this more in Sec.~\ref{sec:discussion}.
Our method, described in Sec.~\ref{sec:method}, extends position embedding $\bm{\psi}_{ij(\bm{n})}$ in Eq.~\ref{eq:attention} to incorporate richer, invariant information beyond distances $\|\bm{p}_{j(\bm{n})} - \bm{p}_i \|$.

\subsection{Frames for SE(3)-invariant structural modeling}
\label{sec:existing_frames}
\textbf{Frame averaging.}
\cite{puny2022frameaveraging} have introduced Frame Averaging (FA) as a general framework to adapt networks to be invariant (or equivariant) to certain symmetries in input data. 
Although FA is originally grounded in group representation theory, we provide a high-level review focused on SE(3)-invariant modeling of 3D point clouds. Given a point cloud as $P$, FA computes a frame,  $F \in \mathcal{F}(P)$,  as a coordinate system inherent to and aligned with $P$ (Fig.~\ref{fig:concept}, left). For instance, $\mathcal{F}$ is principal component analysis (PCA) applied to $P$. Each frame $F$ thus defines a geometric transformation that maps $P$ to a canonical, rotation-invariant representation, $FP$.
However, $\mathcal{F}(P)$ may not uniquely provide a single frame due to algorithmic ambiguities in $\mathcal{F}$ or symmetries in $P$. 
Even in such cases, FA allows us to derive rotation-invariant (\ie, SO(3)-invariant) networks $\bar{f}_{\mathcal{F}}$ from any given networks $f$, by averaging $f$'s outputs over all possible finite frames, as follows:
\begin{equation}
\bar{f}_{\mathcal{F}}(X, P) = \frac{1}{|\mathcal{F}(P)|} \sum_{F \in \mathcal{F}(P)} f(X, FP). \label{eq:frame_averaging}
\end{equation}
The translation invariance is further attained by formulating $f$ with relative positions (\emph{e.g.}, $F\bm{p}_j - F\bm{p}_i$), bringing SE(3) invariance to $\bar{f}_\mathcal{F}$.
FA can powerfully adapt arbitrary networks to be SE(3) invariant without constraining the architectural design. However, it hinders efficiency, as the computation increases with the number of possible frames.
Stochastic FA by \cite{duval23faenet} mitigates this issue by randomly selecting a single frame from $\mathcal{F}(P)$ during training. This scheme enforces networks $f$ to learn the invariance to frame variations, approximately achieving SE(3) invariance.

\textbf{PCA frames.}
\cite{puny2022frameaveraging} originally applied FA for molecules using PCA-based frames, and \cite{duval23faenet} later extended it for crystals by simply treating unit cell structures $P$ as finite-sized point clouds. These PCA frames compute three orthogonal eigenvectors $\{\bm{e}_1, \bm{e}_2, \bm{e}_3\}$ of the covariance matrix for $P$, corresponding to eigenvalues $\lambda_1 \ge \lambda_2 \ge \lambda_3$, as the frame axes: $F = [\bm{e}_1, \bm{e}_2, \bm{e}_3]^T$. Because of the sign ambiguity of the eigenvectors, PCA produces eight frames for O(3)/E(3) invariance and four frames for SO(3)/SE(3) invariance with the restriction of $\det(F)=1$. Although PCA is well-established, it suffers from eigenvalue degeneration for highly symmetric data, such as crystal structures. For example, PCA for cubes produces identity covariance matrices up to a constant scale, whose eigenvectors are arbitrary vectors $\bm{e} \in \mathbb{R}^3$. The crystal frame construction by \cite{duval23faenet} is thus vulnerable to this degeneration issue and, moreover, sensitive to unit-cell variations of the same crystal structure. 

\textbf{Lattice frames.}
\cite{yan2024comformer} have proposed frames based on the lattice vectors of crystals, as similar to \emph{reduced cells} (\ie, uniquely determined minimum cells)~\citep{niggli}. Specifically, their method selects a lattice point,
$\bm{e} = n_1\bm{\ell}_1 + n_2\bm{\ell}_2 + n_3\bm{\ell}_3$, 
with the minimum non-zero norm $\| \bm{e}\|_2$ as first axis $\bm{e}_1$, and selects the second and third smallest ones as axes $\bm{e}_2$ and $\bm{e}_3$ while ensuring $\text{rank}(\bm{e}_1,\bm{e}_2,\bm{e}_3)$ is full. The signs of these axes are adjusted so that the angles between $\bm{e}_1$ and $\bm{e}_2$ and between $\bm{e}_1$ and $\bm{e}_3$ become acute and the coordinate system is right-handed (\ie, $\det{(F)} > 0$).

Notice that these existing frame methods for crystals, specifically PCA and lattice frames, all provide a statically fixed frame for each crystal structure. Also, both rely on unit cell representations (either points $P$ or lattice vectors $L$),
which are rather artificially-introduced crystal descriptions that may not necessarily reflect the physical properties of materials (see Appendix~\ref{appendix:primitive_cell} for more discussion).
These observations motivate us to propose the concept of dynamic frames, as we discuss next.

\section{Dynamic frames}
\label{sec:method}
In the search for effective frames for crystals, we challenge the conventional notion of frames, which implicitly follows the simple premise of representing structures in a canonical manner~\citep{puny2022frameaveraging,duval23faenet,yan2024comformer}. Let us reconsider how frames work in GNNs, whose interatomic message-passing layers are assumed to include the following general operation:
\begin{equation}
    \bm{x}'_i = \sum_{j=1}^{N} \sum_{\bm{n}\in \mathbb{Z}^3} w_{ij(\bm{n})} \bm{f}_{i\gets j(\bm{n})}(\bm{x}_{j(\bm{n})}, \hat{P}). \label{eq:message_passing}
\end{equation}
This equation describes that state $\bm{x}_i$ of each unit-cell atom $i$ is evolved through abstract influences or \emph{messages}, $\bm{f}_{i\gets j(\bm{n})}$, from atoms $j(\bm{n})$ in the crystal structure, with scaling weights $w_{ij(\bm{n})}$. 
In standard GNNs~\citep{xie18cgcnn}, these weights are pre-defined as neighborhood graphs with a cut-off radius. In recent transformer architectures, the weights are determined dynamically via self-attention, with~\citep{yan22matformer,yan2024comformer} or without~\citep{taniai2024crystalformer} relying on an explicit cut-off radius.

The role of frames in Eq.~\ref{eq:message_passing} is to offer, for the design of $\bm{f}_{i \gets j(\bm{n})}$, more informative invariant edge features than distances through frame-projected coordinates $F\hat{P}$. From this perspective, constructing a shared frame for the state updates of all atoms $i$, as done in conventional methods, is not preferable. This is because the frame construction can be influenced even by atoms $j(\bm{n})$ with zero weights in Eq.~\ref{eq:message_passing}.
In other words, when updating the state of atom $i$  in Eq.~\ref{eq:message_passing}, this atom  has its own partial and local view of the entire crystal structure, $\hat{P}$, where weights $w_{ij(\bm{n})}$ act as a mask on the structure.

This interpretation leads to the concept of \emph{dynamic frames}. That is, we define frames locally for each atom $i$ to align with its interatomic interactions dynamically acting on the structure, rather than directly aligning with the structure itself. 
We denote these dynamic atom-wise frames as $F_i$. Each $F_i$ is determined based on the masked view of structure $\hat{P}$ with weights ${w}_{ij(\bm{n})}$, by emphasizing or de-emphasizing the presence of atoms $j(\bm{n})$ with larger or smaller weights. 
Thus, these frames $F_i$ change dynamically depending on target atoms $i$ and also on the layers in a GNN, as shown in Fig.~\ref{fig:concept}.

We hypothesize that dynamically adapting frames for each atom $i$ in each message-passing layer (Eq.~\ref{eq:message_passing}) provides better invariant edge features via projected coordinates $F_i\hat{P}$. We also point out that these frames are defined with the entire crystal structure, $\hat{P}$, reconstructed from $(P, L)$. This fact highlights an advantage of our frames being invariant to unit cell variations within the same structure.

\subsection{Frame definitions}
\label{sec:frame_definition}
We now present several instances of this new family of frames. 
These frames $F_i$ are constructed for each target atom $i$ in each message-passing layer (Eq.~\ref{eq:message_passing}), by using coordinates $\hat{P}$ and weights $w_{ij(\bm{n})}$ of atoms $j(\bm{n})$ in the structure. We typically assume $w_{ij(\bm{n})} \ge 0$, but we can use real-valued weights, for example, by using their absolute values for frame construction. For brevity, we denote $r_{ij(\bm{n})}=\|\bm{p}_{j(\bm{n})}-\bm{p}_i\|_2$ and $\bar{\bm{r}}_{ij(\bm{n})}=(\bm{p}_{j(\bm{n})}-\bm{p}_i)/r_{ij(\bm{n})}$, both derived from $\hat{P}$.

\textbf{Weighted PCA frames.} The first instance of dynamic frames extends the original PCA frames~\citep{puny2022frameaveraging,duval23faenet}. For each target atom $i$ in each message-passing layer, we compute a $3\times3$ weighted covariance matrix, 
$\Sigma_i =\sum_j \sum_{\bm{n}} w_{ij(\bm{n})}\bar{\bm{r}}_{ij(\bm{n})}\bar{\bm{r}}^T_{ij(\bm{n})}$, 
and computes its orthogonal eigenvectors $\{\bm{e}_1, \bm{e}_2, \bm{e}_3 \}$ as the frame axes: $F_i = [\bm{e}_1, \bm{e}_2, \bm{e}_3]^T$. For the sign ambiguity of eigenvectors, we adopt the stochastic FA~\citep{duval23faenet} and generate a single frame by randomly flipping the signs of these vectors while ensuring $\det(F_i) = 1$. However, there remains another possible ambiguity in this weighted PCA scheme owing to eigenvalue degeneration by symmetries\footnote{We confirmed that covariance matrices $\Sigma_i$ computed with a pretrained Crystalformer model suffered from eigenvalue degeneration at two degrees in about 10\% of cases and at three degrees in about 1\% of cases.
These cases cause rotation ambiguities for two or three (all) axes of $F_i$. To mitigate this issue, we add small perturbation noise to $w_{ij(\bm{n})}$ in $\Sigma_i$, which stochastically breaks the symmetries in the structural data and empirically helps to compute non-degenerate eigenvalues and eigenvectors. This scheme is considered a type of stochastic FA.
}.

\textbf{Max frames.} To avoid the degeneration of PCA, we also propose directly selecting atoms $j(\bm{n})$ with large weights $w_{ij(\bm{n})}$ and using their directions $\bar{\bm{r}}_{ij(\bm{n})}$ to determine axes $\{\bm{e}_1, \bm{e}_2, \bm{e}_3 \}$ of $F_i$. Specifically, the first axis, $\bm{e}_1$, is selected as  $\bar{\bm{r}}_{ij(\bm{n})}$ with the maximum weight $w_{ij(\bm{n})}$. For the second axis, we find $\bar{\bm{r}}_{ij(\bm{n})}$ with the maximum adjusted-weight $(1-|\bm{e}_{1} \cdot \bm{r}_{ij(\bm{n})}|)w_{ij(\bm{n})}$, avoiding a direction parallel to $\bm{e}_1$. The selected vector, denoted as $\bar{\bm{r}}_2$, is further orthogonalized using the Gram-Schmidt method as $\hat{\bm{e}}_2 \gets \bar{\bm{r}}_2 - (\bm{e}_1 \cdot \bar{\bm{r}}_2) \bm{e}_1$, and normalized to a unit vector as $\bm{e}_2 \gets \hat{\bm{e}}_2 / \|\hat{\bm{e}}_2\|_2$. Finally, the third axis is simply obtained as $\bm{e}_3 \gets \bm{e}_1\times \bm{e}_2$, which ensures the orthogonality and $\det(F_i) = 1$. 
In this process, multiple atoms may have the same weight. For this ambiguity, we add small perturbation noise to each weight $w_{ij(\bm{n})}$, resulting in randomly selecting a single frame from possible ones. This perturbation scheme is considered a type of stochastic FA~\citep{duval23faenet} outlined in Sec.~\ref{sec:existing_frames}.

Since these frame construction processes are not stably differentiable, we omit the computation of the gradients from frames $F_i$ to weights $w_{ij(\bm{n})}$ during training\footnote{The gradients of the eigenvectors in PCA become numerically unstable when the eigenvalues are degenerate, as the gradients depend on the computation of $1/(\lambda_i - \lambda_j)$ for ${i \neq j}$. Also, the max-frame procedure is not differentiable due to the use of argmax operations. Although we tried approximating the gradients of argmax, for example, by using a straight-through estimator technique or temperature annealing of softmax, simply ignoring the frame gradients gave the best results.}.
Still, weights $w_{ij(\bm{n})}$ receive gradients from $\bm{x}'$ in Eq.~\ref{eq:message_passing} to learn their main function: allowing or blocking messages $\bm{f}_{i \gets j(\bm{n})}$ from $j(\bm{n})$ to $i$. Therefore, we can train a network with dynamic frame construction without using frame gradients.

\subsection{CrystalFramer architecture}
\label{sec:architecture}
\begin{figure}[t]
\centering
\includegraphics[width=1.0\textwidth]{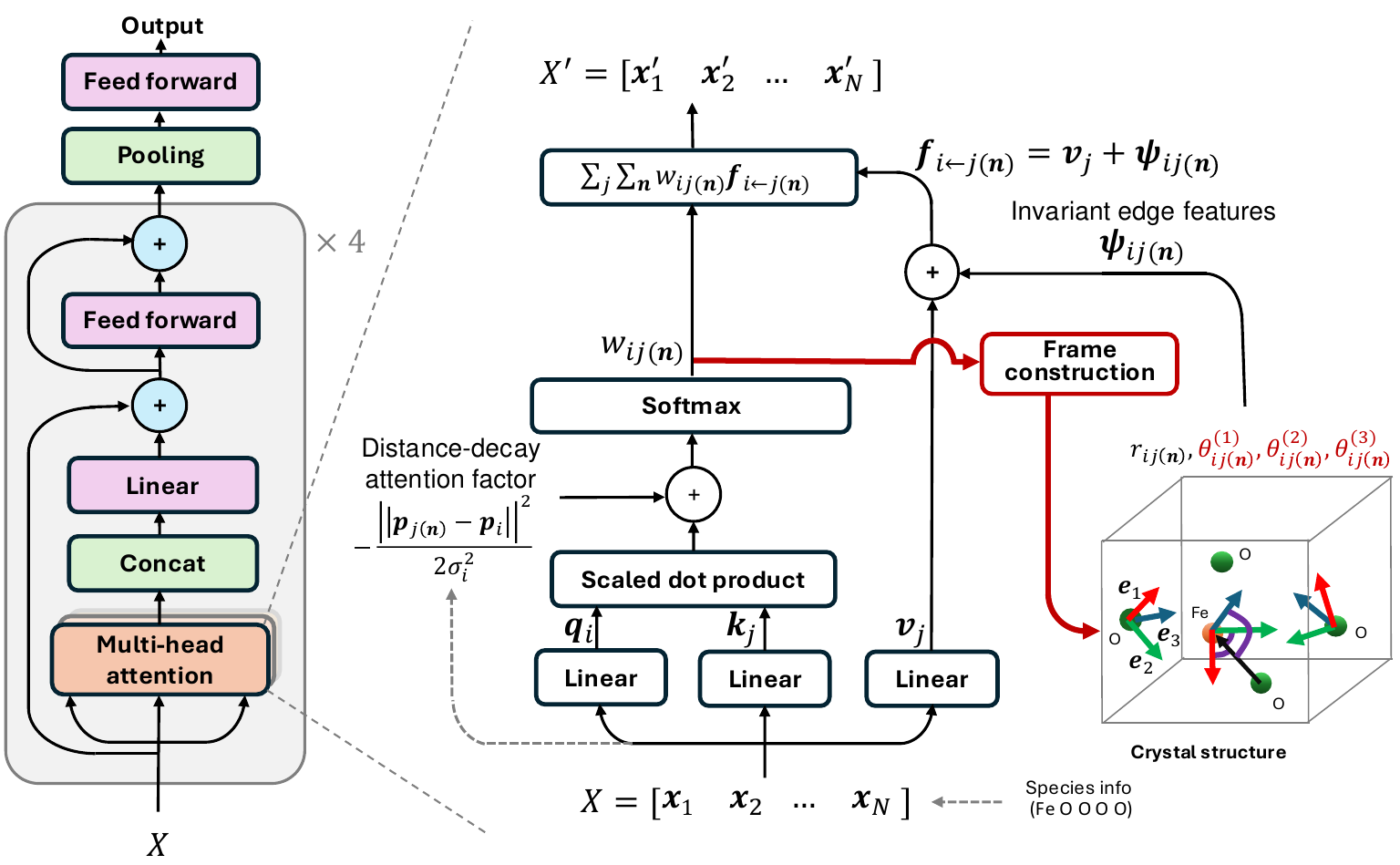}
\caption{\textbf{CrystalFramer architecture.} Dynamic frame construction and frame-based invariant edge features (highlighted in red) are introduced to a transformer for crystals~\citep{taniai2024crystalformer}.}
\label{fig:attention_layer}
\end{figure}

We demonstrate the proposed concept using  Crystalformer~\citep{taniai2024crystalformer} as the baseline, as mentioned in Sec.~\ref{sec:crystalformer}, and develop a new architecture, CrystalFramer (Fig.~\ref{fig:attention_layer}).

We here regard Eq.~\ref{eq:attention} as Eq.~\ref{eq:message_passing}. Thus, we regard the softmax self-attention weights (\ie, exponential weights normalized by $Z_i$ in Eq.~\ref{eq:attention}) as dynamic scaling weights $w_{ij(\bm{n})}$ in each message-passing layer (Eq.~\ref{eq:message_passing}). 
Likewise, we regard the position-augmented value vectors, $\bm{v}_j+\bm{\psi}_{ij(\bm{n})}$, as messages $\bm{f}_{i \gets j(\bm{n})}$.
During the state update of each atom $\bm{x}_i$ using Eq.~\ref{eq:attention}, we first compute the attention weights as $w_{ij(\bm{n})}$. Then, we dynamically construct a local frame for each atom $i$ as matrix $F_i$, by following one of the procedures outlined in Sec.~\ref{sec:frame_definition}. Finally, we compute $\bm{\psi}_{ij(\bm{n})}$ using $F_i$ and perform Eq.~\ref{eq:attention}. Below, we explain how to derive invariant edge features $\bm{\psi}_{ij(\bm{n})}$, given that frame $F_i$ is determined.

\textbf{Invariant edge features using a dynamic frame.} 
For invariant edge feature $\psi_{ij(\bm{n})}$, Crystalformer originally uses linearly projected Gaussian basis functions (GBFs) to encode distance $r_{ij(\bm{n})}$. Specifically, GBFs are defined as a mapping from a scalar to a vector of pre-defined dimension $D$, $\bm{b}(x) = [b_1, b_2, \cdots, b_D]^T$, whose $k$-th component is computed as a Gaussian given by
\begin{equation}
b_k(x;\mu_k, \sigma_k)=\exp{\left(-(x-\mu_k)^2/2\sigma_k^2\right)}.
\end{equation}
Here, $\mu_k$ and $\sigma_k$ are pre-defined  as $\mu_k = \mu_{\text{min}} + (k-1)(\mu_{\text{max}} - \mu_{\text{min}})/(D-1)$ and $\sigma_k = s(\mu_{\text{max}}-\mu_{\text{min}})/(D-1)$, where $\{\mu_{\text{max}}, \mu_{\text{min}}, s, D\}$ are hyperparameters. Intuitively, $\bm{b}(x)$ encodes scalar $x$ into a soft one-hot vector using $D$ Gaussians uniformly distributed between $\mu_\text{min}$ and $\mu_\text{max}$. The widths of these Gaussians, $\sigma_k$, are proportional to the interval distance, controlled by scaling factor $s$.

We retain their distance-based edge feature and further add frame-based edge features to $\bm{\psi}_{ij(\bm{n})}$. Specifically, following existing work~\citep{yan2024comformer}, we represent direction vector $\bar{\bm{r}}_{ij(\bm{n})}$ invariantly by projecting it onto the frame coordinate system, as $\bm{\theta}_{ij(\bm{n})} = F_i \bar{\bm{r}}_{ij(\bm{n})}$. Its $k$-th component is calculated as $\bm{e}_k\cdot \bar{\bm{r}}_{ij(\bm{n})}$, the cosine of the angle between $k$-th frame axis $\bm{e}_k$ and direction $\bar{\bm{r}}_{ij(\bm{n})}$. Each component is then converted to a vector using GBFs. By combining the distance-based and three angle-based features via linear projections, we obtain our geometric relative position encoding:
\begin{equation}
\label{eq:value_pos_encoding}
\bm{\psi}_{ij(\bm{n})} = \lambda \left( c_\text{dist} W_0 \bm{b}_\text{dist}\left(r_{ij(\bm{n})}\right) + c_\text{angl} \sum_{k=1,2,3} W_k\bm{b}_\text{angl}\left(\theta_{ij(\bm{n})}^{(k)}\right) \right).
\end{equation}
This $\bm{\psi}_{ij(\bm{n})}$ as a whole essentially encodes the 3D relative position vector: $\bm{r}_{ij(\bm{n})} = \bm{p}_{j(\bm{n})} - \bm{p}_i$. Furthermore, the angle components of $\bm{\psi}_{ij(\bm{n})}$ can be interpreted as encoding the deviations of $\bm{r}_{ij(\bm{n})}$ from the three primary directions of interatomic interactions (\ie,  $\bm{e}_1,\bm{e}_2,\bm{e}_3$) around target atom~$i$. 
Here, four weight matrices $\{W_0, W_1, W_2, W_3\}$ are trainable parameters provided per layer, and coefficients $\{\lambda, c_\text{dist}, c_\text{angl}\}$ are hyperparameters. We also use two types of GBFs, $\bm{b}_{\text{dist}}$ and $\bm{b}_{\text{angl}}$, with different hyperparameters for the distance and angles. 

\textbf{Default and lightweight configurations.}
The default configuration is designed to introduce minimal modifications to the baseline.
With $\lambda = 1.0$, we set $\{c_\text{dist}, \mu_{\text{min}}, \mu_{\text{max}}, s, D\}$ to $\{1.0, \frac{14.0}{64}\text{\AA}, 14.0 \text{\AA}, 1.0, 64\}$ for the distance-based term, maintaining the same definition as in the baseline~\citep{taniai2024crystalformer}.
For the angular term, we set $\{c_\text{angl}, \mu_{\text{min}}, \mu_{\text{max}}, s, D\}$ to $\{1.0, -1.0, 1.0, 4.0, 64\}$, using the same coefficient and dimension as the distance-based term, with a cosine range of $[-1.0, 1.0]$ and a tuned width scale~$s$. This default configuration was used in most experiments in Sec.~\ref{sec:experiments}.
With this setup, $\bm{b}_\text{angl}(x)$ employs largely overlapping Gaussian bases (\ie, $s = 4.0$), resulting in a smooth variation along $x$. This suggests that $\bm{b}_\text{angl}(x)$ can be well-approximated using fewer Gaussian bases (\ie, a lower $D$) to improve efficiency.
By reducing $D$ to $D'$, comparable Gaussian widths and output value ranges can be obtained by adjusting the scale to $s' = sD'/D$ and the coefficient to $c' = cD/D'$. Applying this transformation, we derived a lightweight configuration by modifying the angular term’s $\{c_\text{angl}, s, D\}$ to $\{4.0, 1.0, 16\}$ and further tuning $\lambda$ to $1.5$. This configuration reduces training time, as shown in Sec.~\ref{sec:efficiency}.

\textbf{Overall architecture.} 
The proposed network precisely follows the Crystalformer architecture~\citep{taniai2024crystalformer} as shown in Fig.~\ref{fig:attention_layer}, with the addition of frame construction (Sec.~\ref{sec:frame_definition}) and angular edge features (Eq.~\ref{eq:value_pos_encoding}), highlighted in the figure. As we will see in Sec.~\ref{sec:experiments}, these simple extensions lead to drastic performance improvements over the baseline.
Below, we summarize the key design aspects of the network. The overall architecture consists of an input atom-embedding layer, a stack of four self-attention blocks, global mean pooling, and a final feed-forward network with two linear layers.
The self-attention blocks adopt a normalization-free architecture (left part of Fig.~\ref{fig:attention_layer}) by \cite{huang20improving} for enhancing training stability. 
The infinite summation in self-attention (Eq.~\ref{eq:attention}) is computed convergently and efficiently. Specifically, we adaptively determine the range of unit-cell shifts $\bm{n}$ to sufficiently cover the neighbor radius of $3.5\sigma_i$, based on dynamic Gaussian tail-length $\sigma_i$. We also employ multi-head self-attention as in the original transformer~\citep{vaswani17transformer}, with eight heads. Frames are constructed per unit-cell atom, per head, and per layer. For further architectural details, please refer to the original work~\citep{taniai2024crystalformer}.

\section{Related work}
The notion of invariant structural modeling encompasses various invariance properties. The most basic one is invariance to the permutation of data-point indices $i$ in structural data (\ie, $A$ and $P$ in Sec.~\ref{sec:crystal_structure}). It was first addressed by PointNet~\citep{qi17pointnet} and DeepSets~\citep{zaheer17deepsets} and is now inherited by GNNs and transformers. The ML community has subsequently focused on invariance under geometric transformations, such as rotations with or without translations (\ie, SO(3)/O(3) or SE(3)/E(3) invariance). In particular, the periodicity of crystals introduces more complex invariance notions, such as periodic SE(3) invariance~\citep{yan22matformer}, requiring invariance to unit-cell variations within the same crystal structure. These geometric invariance properties have been explored in three primary approaches: 1) invariant features, 2) equivariant features, and 3) frames. We briefly review these approaches, focusing primarily on crystal structures.

\textbf{Invariant features.} The most straightforward approach is to rely entirely on inherently invariant geometric quantities, such as the lengths of relative position vectors, throughout a model~\citep{xie18cgcnn,chen2019megnet,ying21graphormer,yan22matformer,taniai2024crystalformer}. However, such distance-based GNNs and transformers have limited expressibility~\citep{pozdnyakov22incomplete}. Thus, recent studies have explored more advanced geometric features, such as 
the angles between triplets in 3-body interactions~\citep{park20icgcnn,choudhary21alignn,chen22m3gnet}, though at the cost of increased computational complexity. More recently, PotNet~\citep{lin2023potnet} employed the periodic summation of pre-defined interatomic scalar potentials 
as more physically informed invariant edge features, compared to distances.

\textbf{Equivariant features.} Equivariant networks, grounded in group representation theory, form another active research area in 3D structural modeling and include invariant networks as special cases. 
While we refer readers to recent surveys~\citep{Gerken2023,duval2024hitchhikersguidegeometricgnns,han2024surveygeometricgraphneural} for more comprehensive reviews, the initial approach using GNNs for 3D point clouds and atomic systems was proposed by \cite{thomas2018tfn}. Subsequently, this approach has been extended, for example, to introduce improved nonlinearities~\citep{Batzner2022nequip,brandstetter2022segnn}, attention mechanisms~\citep{fuchs2020se3transformer}, or greater efficiency~\citep{liao2023equiformer,liao2024equiformerv2} in molecular structure modeling. 
Essentially, these methods use spherical harmonic representations of unit direction vectors $\bar{\bm{r}}_{ij}$ as rotation-equivariant edge features, which are then equivariantly transformed through specially designed networks. 
These equivariant features form type-$L$ vectors in a pyramidal shape, where type-0 features encode invariant scalars and type-1 features represent 3D equivariant vectors, such as forces. 
However, these networks are constrained by limited nonlinearity forms and the increasing computational complexity required to incorporate higher-frequency components in higher degrees $L$. Due to these limitations, their application in crystals is more limited compared to molecules. For example, eComFormer~\citep{yan2024comformer} has exploited equivariant features in part within each message-passing block for invariant crystal property prediction.

\textbf{Frames.} As explained in Sec.~\ref{sec:existing_frames}, \cite{puny2022frameaveraging} introduced FA and applied PCA frames for molecules. \cite{duval23faenet} further extended this work in two ways: by proposing stochastic FA to improve efficiency and by applying PCA frames to crystals by treating their unit-cell structures $P$ as finite point clouds. 
\cite{cheng21geocgnn} used plane waves in crystal structures as invariant positional features, implicitly employing reciprocal lattice vectors as a frame. Similarly, \cite{yan2024comformer} proposed iComFormer, using transformed lattice vectors with reduced ambiguities as a frame. 
\cite{lin2024minimalfa} proposed minimal FA, an efficient FA method ensuring exact invariance and equivariance.

Our work contributes to this line of research on frame-based invariant networks, providing a new perspective on the previous notion of frames through the introduction of dynamic frames. 
While several local frame methods exist in the molecular modeling literature~\citep{du2022localframe,du2023localframetrans,pozdnyakov2023ensemble}, they do not incorporate dynamic frames as we do (see Appendix~\ref{appendix:localframe} for a comparative discussion).
We integrate these dynamic frames into a simple distance-based transformer model for crystals~\citep{taniai2024crystalformer} to boost its expressive power.

\section{Experiments}
\label{sec:experiments}
To validate the effectiveness of the proposed dynamic frames, we conducted extensive experiments on crystal property prediction. We compared our method with conventional PCA frames~\citep{duval23faenet}, lattice frames~\citep{yan2024comformer}, and other state-of-the-art architectures for crystals.

\textbf{Datasets.}
We use three datasets: JARVIS (55,723 materials), MP (69,239 materials), and OQMD (817,636 materials), using snapshots available through a Python package ({jarvis-tools}). These datasets provide several material properties, such as formation energy and bandgap, simulated by DFT calculations.
For further dataset descriptions, see Appendix~\ref{appendix:dataset}. 
\cite{choudhary21alignn} and \cite{yan22matformer} evaluated many methods on the JARVIS and MP datasets using consistent data splits. Following these and later studies~\citep{lin2023potnet,yan2024comformer,taniai2024crystalformer}, we use the same data splits and cite their reported scores to reduce computational burden.
Unlike these studies, we also use the much larger-scale OQMD dataset to assess scalability.

\textbf{Training settings.}
To assess the pure effects of introducing the frames, we precisely follow the training settings of the baseline method, Crystalformer~\citep{taniai2024crystalformer}, with only one modification. We have increased the number of training epochs to account for the increased complexity of our edge feature design (\ie, our method takes longer to converge, but reduces validation losses more rapidly). Specifically, for the JARVIS dataset, we train our model from scratch by optimizing the mean absolute loss function using Adam~\citep{kingma14adam} for a total of 2000 epochs, while enabling the frames from the beginning. 
A summary of detailed training settings, including the number of epochs, batch size, and learning rate for the three datasets, can be found in Appendix~\ref{appendix:detail_train_param}.

\subsection{Crystal property prediction}
\label{sec:property_prediction}

\robustify\bfseries
\begin{table}[p]
\vskip -2.5\baselineskip
\caption{\textbf{Property prediction results on the JARVIS dataset.} Accuracies are in mean absolute error. The sizes of training, validation, and test splits are listed under each property name. \textbf{Bold} indicates the best results, \underline{underline} the second best.}
\label{table:results_jarvis}
\scriptsize
  \centering
  \vskip 0.5mm
  \begin{tabular}{lS[table-format=1.4]S[table-format=1.4]S[table-format=1.3]S[table-format=1.3]S[table-format=1.4]}
    \toprule
    & {Form. energy} & {Total energy} & {Bandgap (OPT)}  & {Bandgap (MBJ)} & {E hull} \\
    & {\tiny 44578\,/\,5572\,/\,5572} & {\tiny 44578\,/\,5572\,/\,5572} & {\tiny 44578\,/\,5572\,/\,5572} & {\tiny 14537\,/\,1817\,/\,1817} & {\tiny 44296\,/\,5537\,/\,5537} \\
    \cmidrule(r){2-6}
    Method & {\tiny eV/atom}  &  {\tiny eV/atom} & {\tiny eV} & {\tiny eV} & {\tiny eV}    \\
    \midrule
    CGCNN~\citep{xie18cgcnn}  & 0.063 &  0.078 &  0.20 & 0.41& 0.17   \\
    SchNet~\citep{schutt18schnet} & 0.045 &   0.047 & 0.19 & 0.43 & 0.14   \\
    MEGNet~\citep{chen2019megnet}  & 0.047 &    0.058 & 0.145 & 0.34 & 0.084  \\
    GATGNN~\citep{louis20gatgnn}  & 0.047 &   0.056 & 0.17 & 0.51 & 0.12              \\
    M3GNet~\cite{chen22m3gnet} & 0.039 &	0.041 &	0.145 &	0.362 &	0.095 \\
    ALIGNN~\citep{choudhary21alignn} & 0.0331 &  0.037 & 0.142 & 0.31 & 0.076   \\
    Matformer~\citep{yan22matformer} & 0.0325 & 0.035 & 0.137 & 0.30 & 0.064    \\
    PotNet~\citep{lin2023potnet} &  0.0294 & 0.032 & 0.127 & 0.27 & 0.055 \\
    eComFormer~\citep{yan2024comformer} & 0.0284 & 0.032 & 0.124 & 0.28 & \bfseries 0.044 \\
    iComFormer~\citep{yan2024comformer} & \uline{\tablenum[table-format=1.4]{0.0272}} & \uline{\tablenum[table-format=1.4]{0.0288}} & \uline{\tablenum[table-format=1.3]{0.122}} & \uline{\tablenum[table-format=1.3]{0.26}} & 0.047 \\
    \midrule
    Crystalformer~\citep{taniai2024crystalformer} & 0.0306 & 0.0320 & 0.128 & 0.274 & \uline{0.0463}   \\
    --- w/ PCA frames~\citep{duval23faenet} & 0.0325 & 0.0334 & 0.144 & 0.292 & 0.0568 \\
    --- w/ lattice frames~\citep{yan2024comformer} & 0.0302 & 0.0323 & 0.125 & 0.274 & 0.0531 \\
    --- w/ static local frames & 0.0285 & 0.0292 & \underline{0.122} & \underline{0.261} & \bfseries 0.0444 \\
    --- w/ \textbf{weighted PCA frames} (proposed) & 0.0287 & 0.0305 & 0.126 & 0.279 & \bfseries 0.0444 \\
    --- w/ \textbf{max frames} (proposed) & \bfseries 0.0263 & \bfseries 0.0279 & \bfseries 0.117 & \bfseries 0.242 & 0.0471 \\
    \midrule 
    \midrule 
    CrystalFramer (default) & \bfseries 0.0263 & \bfseries 0.0279 & \bfseries 0.117 & \bfseries 0.242 & 0.0471 \\
    CrystalFramer (lightweight) & 0.0268	& \bfseries 0.0279	& \bfseries 0.117	& 0.262 & \bfseries 0.0467 \\
    \bottomrule
  \end{tabular}
\caption{\textbf{Property prediction results on the MP dataset.}
  }
\label{table:results_mp}
\scriptsize
  \centering
  \vskip 0.5mm
  \begin{tabular}{lS[table-format=1.4]cS[table-format=1.4]S[table-format=1.4]}
    \toprule
    & {Formation energy} & {Bandgap} & {Bulk modulus} & {Shear modulus} \\
     & \tiny{60000\,/\,5000\,/\,4239} & \tiny{60000\,/\,5000\,/\,4239} & \tiny{4664\,/\,393\,/\,393} & \tiny{4664\,/\,392\,/\,393} \\
    \cmidrule(r){2-5}
    Method & {\tiny{eV/atom}}  &  \tiny{eV} &   {\tiny{log(GPa)}} & {\tiny{log(GPa)}}  \\
    \midrule
    CGCNN & 0.031 & 0.292  & 0.047 &0.077 \\
    SchNet & 0.033 & 0.345 & 0.066 & 0.099 \\
    MEGNet & 0.030 & 0.307 & 0.060 & 0.099 \\
    GATGNN & 0.033 & 0.280 & 0.045 & 0.075 \\
    M3GNet & 0.024 & 0.247 & 0.050 & 0.087 \\
    ALIGNN & 0.022 & 0.218 & 0.051 & 0.078 \\
    Matformer & 0.021 & 0.211 & 0.043 & 0.073 \\
    PotNet & 0.0188 & 0.204 & 0.040 & \uline{\tablenum[table-format=1.4]{0.065}} \\
    eComFormer & 0.0182 & 0.202 & 0.0417 & 0.0729 \\
    iComFormer & 0.0183 & 0.193 & 0.0380 & \textbf{0.0637} \\
    \midrule
    {Crystalformer} & 0.0186 & 0.198 & 0.0377 & 0.0689 \\
    --- w/ PCA frames & 0.0197 & 0.217 & 0.0424 & 0.0719 \\
    --- w/ lattice frames & 0.0194 & 0.212 & 0.0389 & 0.0720 \\
    --- w/ static local frames & \underline{0.0178} & \underline{0.191} & \underline{0.0354} & 0.0708 \\
    --- w/ \textbf{weighted PCA frames} (proposed) & 0.0197 & 0.214 & 0.0423 & 0.0715 \\
    --- w/ \textbf{max frames} (proposed) & \textbf{0.0172} & \textbf{0.185} & \textbf{0.0338} & 0.0677 \\
    \midrule
    \midrule
    CrystalFramer (default) & \textbf{0.0172} & \textbf{0.185} & \textbf{0.0338} & 0.0677 \\
    CrystalFramer (lightweight) & 0.0176 & 0.191 & 0.0341 & \textbf{0.0654}
  \\
    \bottomrule
  \end{tabular}\\
\caption{\textbf{Property prediction results on the OQMD dataset.} }
\label{table:results_oqmd}
\scriptsize
  \centering
  \vskip 0.5mm
    \begin{tabular}{lcccc}
    \toprule
    & \# Blocks & {Form. energy (eV/atom)} & {Bandgap (eV)} & {E hull (eV/atom)} \\
Method   &    & \tiny{654108\,/\,81763\,/\,81763} & \tiny{653388\,/\,81673\,/\,81673} & \tiny{654108\,/\,81763\,/\,81763}  \\
    \midrule
    {Crystalformer} & 4 & 0.02115 & {0.06028} & 0.06759  \\
    \textbf{CrystalFramer} (default) & 4 &  \underline{0.01871} & \underline{0.05805} & \bfseries{0.06607}  \\
    \textbf{CrystalFramer} (lightweight) & 4 & \textbf{0.01813} & \bfseries 0.05773 & \underline{0.06672} \\
    \midrule
    \midrule
    {Crystalformer} & 8 & 0.02104 &	0.05986	& 0.06690  \\
    \textbf{CrystalFramer} (default) & 8 &  \underline{0.01778} & \underline{0.05785} & \underline{0.06454}  \\
    \textbf{CrystalFramer} (lightweight) & 8 & \textbf{0.01731} & \bfseries 0.05142 & \bfseries 0.06403 \\
    \bottomrule
  \end{tabular}\\
%
\caption{\textbf{Efficiency comparison.} Per-epoch training time includes validation, and per-material test time includes preprocessing, such as graph construction. The runtimes are evaluated for the formation energy prediction in the JARVIS dataset using a single NVIDIA A6000 GPU with 48GB VRAM.}
\label{table:efficiency}
\scriptsize
\centering
\vskip 0.5mm
\begin{tabular}{lccccc}
    \toprule
    Model &  Arch. type & Time/epoch & Test/mater. & {\ttfamily\#}Params. &  {\ttfamily\#}Params./block \\
    \midrule
 PotNet     & GNN  & 43 s & 313 ms  & 1.8 M & 527 K \\
 Matformer  & Transformer  & 60 s  & 20.4 ms & 2.9 M & 544 K \\
 iComFormer  & Transformer  & 59 s & 54.8 ms & 5.0 M & 855 K \\
 Crystalformer & Transformer & 32 s & 6.6 ms & 853 K & 206 K \\
 \textbf{CrystalFramer} (default) & Transformer & 74 s & 16.8 ms & 952 K &  231 K \\
 \textbf{CrystalFramer} (lightweight) & Transformer & 43 s & 15.2 ms  & 878 K & 212 K  \\
  \bottomrule
\end{tabular}
\end{table}

We first discuss the performance of our method with the default configuration, followed by an analysis of the lightweight version in Sec.~\ref{sec:efficiency}.

In the top and middle parts of Tables~\ref{table:results_jarvis} and ~\ref{table:results_mp}, we extensively compare the mean absolute errors of the proposed and existing methods for the JARVIS (5 tasks) and MP (4 tasks) datasets. 
Overall, our method with max frames outperforms others in most tasks, significantly enhancing the performance of the baseline Crystalformer model. 
It is important to note that the current state-of-the-art, ComFormer, uses finely-tuned hyperparameters (\eg, learning rate, loss function, number of layers, graph structure) for each individual task, whereas we simply adjust the number of epochs and batch size for each dataset.

Table~\ref{table:results_oqmd} shows that the superiority of our method over the baseline never fades even when feeding the much larger OQMD dataset. Here, we compare Crystalformer and the proposed CrystalFramer model with max frames in two settings, employing either four or eight self-attention blocks.
In both settings, our method consistently outperforms the baseline. We also observe that increasing the number of self-attention blocks improves the performance of both models. Exploring even larger models is left for future work.

In the middle parts of Tables~\ref{table:results_jarvis} and \ref{table:results_mp}, our weighted PCA frame method shows relatively limited improvements (see Sec.~\ref{sec:visual_analysis} for a detailed discussion). Nevertheless, it outperforms its conventional counterpart using PCA frames. 
Additionally, we evaluated a variant using static local frames. These frames are similar to max frames but constructed with static weights, $w_{ij(\bm{n})} = \exp\big(-r_{ij(\bm{n})}^2\big)$. As a result, these static local frames rely only on the distances to neighbors and do not account for dynamic self-attention weights. The max frame method outperforms this static counterpart in most tasks.
These results successfully validate the effectiveness of our concept of dynamic frames.

\subsection{Efficiency comparison}
\label{sec:efficiency}
Table~\ref{table:efficiency} compares the model efficiencies of several top-performing architectures. Notably, despite the superior performance of the proposed method, it requires significantly fewer parameters than PotNet, Matformer, and iComFormer. Compared to Crystalformer, our method with the default configuration introduces only a small overhead of about 100K parameters owing to projection matrices $\{W_1, W_2, W_3\}$ in Eq.~\ref{eq:value_pos_encoding}.
Given the performance gains shown in Tables~\ref{table:results_jarvis}--\ref{table:results_oqmd}, this high cost-performance ratio also highlights the effectiveness of our dynamic frame feature design.
In terms of runtime, the test time is faster than PotNet, Matformer, and iComFormer, which are hindered by relatively heavy data preprocessing. However, compared to Crystalformer, the training and test times are more than double, mainly due to the increased computation cost of $\sum_{\bm{n}} w_{ij(\bm{n})}\bm{\psi}_{ij(\bm{n})}$. As noted by \cite{taniai2024crystalformer}, this term is the primary bottleneck in Crystalformer and also in our model.

To alleviate this bottleneck, we adopt the lightweight configuration described in Sec.~\ref{sec:architecture}. 
Table~\ref{table:efficiency} shows that this configuration reduces training time by approximately 42\%, while maintaining comparable accuracy on the JARVIS and MP datasets (Tables~\ref{table:results_jarvis} and \ref{table:results_mp}) and achieving better accuracy on the OQMD dataset (Table~\ref{table:results_oqmd}).
To further accelerate training, it may be possible to pre-train the model without frames, allowing it to efficiently learn attention weights first. 

In terms of computational complexity, our model is based on a fully connected attention mechanism, and thus has a computational complexity of $O(N^2)$, where $N$ is the number of atoms per unit cell. In Appendix~\ref{appendix:scalability}, we discuss the scalability for large structures and explore possible extensions to improve efficiency in such cases.

\section{Discussion and limitations}
\label{sec:discussion}

\subsection{Visual analysis}
\label{sec:visual_analysis}

\begin{figure}[t]
\centering
\begin{flushleft}\textbf{(a)}\end{flushleft}
\vskip -1.3\baselineskip
\hfill\includegraphics[width=0.95\textwidth]{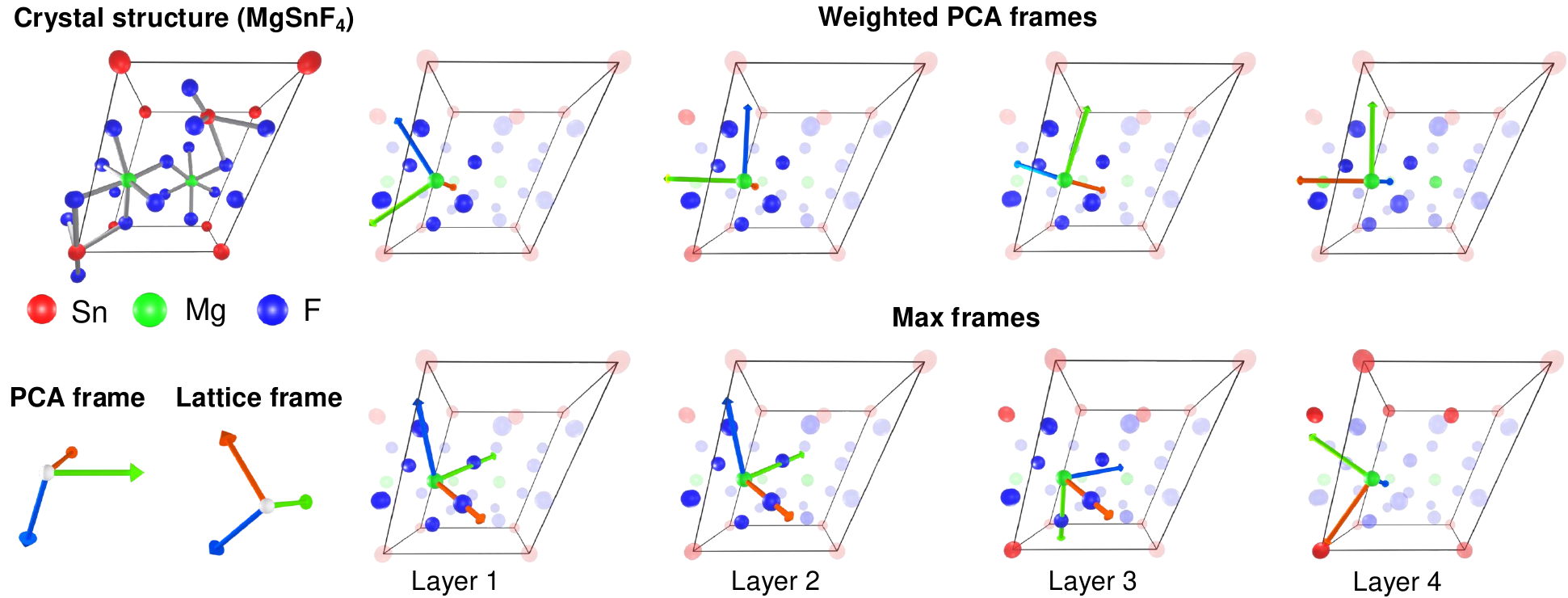}\\
\vskip 5mm
\begin{flushleft}\textbf{(b)}\end{flushleft}
\vskip -1.45\baselineskip
\hspace{1mm}\includegraphics[width=0.95\textwidth]{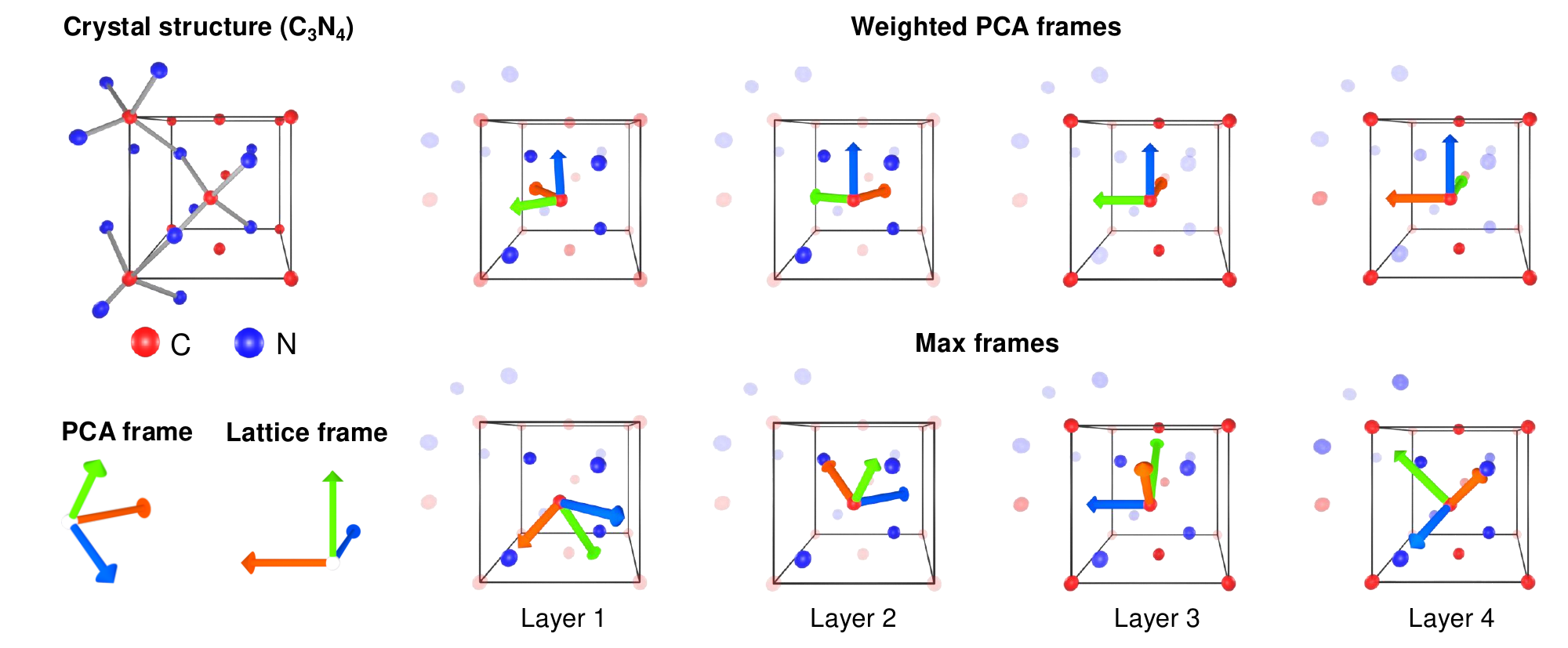}
\caption{\textbf{Frame visualizations.} Conventional PCA and lattice frames provide a global coordinate system based solely on the structure.
The proposed dynamic frames extract different structural information for each atom and layer using dynamic attention weights, shown as varying transparency.}
\label{fig:frame_comparison}
\end{figure}

Figure~\ref{fig:frame_comparison} displays four types of frames generated for two test material (JVASP-30609 and JVASP-85272) in the JARVIS formation energy prediction task. While PCA~\citep{duval23faenet} and lattice~\citep{yan2024comformer} frames are static, the proposed weighted PCA and max frames exhibit dynamic behavior based on learned attention weights. Below, we provided a detailed analysis on our dynamic frames, discussing possible reasons for the superior performance of max frames.

\textbf{Max frames.}
In Fig.~\ref{fig:frame_comparison}a, our max frames capture distinct local motifs, such as octahedra with a green central magnesium atom surrounded by blue fluorine atoms, and tetrahedra with a red central tin atom surrounded by blue magnesium atoms. These local structures are common and often distorted, as seen in this example.
The ability to capture these local structures and measure distortions via relative positions may contribute to the high performance observed. 
In Fig.~\ref{fig:frame_comparison}b, the structure consists of red carbon atoms and blue nitrogen atoms, forming a tetrahedral structure. Both types of dynamic frames initially attend to the central tetrahedral structure in the first two layers, and then increase the attention to relatively distant red atoms in the subsequent layers. However, the max frames capture these structures more clearly than the weighted PCA frames, as observed in the first example.

\textbf{Weighted PCA frames.} In both examples, the weighted PCA frames do not seem to capture the local structure as well as the max frames. This is because all attention weights, even small ones, can influence the composition of the weighted PCA frames. In other words, the weighted PCA frames look at the structure over a broader area, while the max frames focus on relatively close neighbors.
This difference seems to have a positive effect on the max frames and a negative effect on the weighted PCA frames in most tasks, except for the E hull in the MP dataset (Table~\ref{table:results_mp}).
For the E hull prediction, \cite{taniai2024crystalformer} suggest that the inclusion of long-range interatomic interactions is a critical factor. This implication can reasonably explain the better performance of the weighted PCA frames for the E hull. That is, the weighted PCA frames emphasize distant atoms and help deliver more meaningful messages from these distant atoms that are important for the E hull prediction.

\textbf{Other observations.} We have also noticed a general tendency for our models to attend to close neighbors in shallow layers and relatively distant neighbors in deeper layers. This tendency is reasonable. Since the states of atoms are initialized as symbolic atomic species without rich information, they must gather information about their surroundings in shallow layers to configure their states. In deeper layers, these atoms become ready to engage in complex interactions with selected distant atoms.

In Appendix~\ref{appendix:evolution}, we further provide an examination of frame evolution during training. This analysis shows that max frames converge faster during training due to the discrete nature of their construction. While this characteristic may contribute to the superior performance of max frames, it introduces noticeable discontinuities to the model and may limit generalization to out-of-domain data, as discussed in Appendix~\ref{appendix:continuity}.

\subsection{Baseline choice} This study adopts Crystalformer~\citep{taniai2024crystalformer} for demonstration, since its standard multi-head softmax attention is well-suited for dynamic frames. Other crystal transformers~\citep{yan22matformer,yan2024comformer} use distinct channel-wise sigmoid attention, akin to maximally multi-headed attention. Since frames and angular features are computed per atom, per head, and per layer, such channel-wise attention is not preferable. On the other hand, Crystalformer with Eq.~\ref{eq:attention} can be interpreted as the original fully-connected self-attention~\citep{vaswani17transformer}, $\bm{x}_i' = Z_i^{-1}\sum_j \exp(\bm{q}_i^T \bm{v}_j/\sqrt{d_K} + \phi_{ij})(\bm{v}_j + \bm{\psi}_{ij})$, with two straightforward extensions: 1) relative position representations via $\phi_{ij}$ and $\bm{\psi}_{ij}$~\citep{shaw18relative} and 2) duplication of each atom $j$ as $j(\bm{n})$ using $\sum_{\bm{n}}$ for crystal periodicity.  Given the practicability and versatility of the original transformer architecture across diverse domains~\citep{lin2021surveytransformers}, our demonstration serves as a foundation for transformer-based crystal encoders using dynamic frames.

\subsection{Equivariant prediction} 
While this study focuses on SE(3) invariance, \cite{puny2022frameaveraging} extended FA to predict equivariant quantities, such as force vectors, by applying inverse mapping $F^{-1}$ to $f(X, FP)$ in Eq.~\ref{eq:frame_averaging} prior to averaging. In our case, one potential equivariant extension would thus invariantly output atom-wise geometric quantities $\bm{u}_i$ from $\bm{x}_i'$ (\eg, via $\bm{u}_i = W\bm{x}_i'$) and subsequently apply the inverse mapping as $F_i^{-1} \bm{u}_i$. Another approach, similar to recent work~\citep{shi2023graphormer3d}, could tie the outputs equivariantly to the input structure, for instance, by computing $\bm{u}_i = \sum_j \sum_{\bm{n}} w_{ij(\bm{n})} \bm{r}_{ij(\bm{n})}$. These equivariant extensions would enable predictions of forces and relaxed structures, which are crucial for surface property analysis~\citep{chanussot2020oc20,tran2022oc22,bihani2024egraff}. Detailed investigations into these equivariant extensions are left for future work.

\subsection{Application to molecules}
Transformers for molecular structures have been developed~\citep{ying21graphormer, wang2023geoformer,shi2023graphormer3d, liu2024gradformer}, and our dynamic frames could potentially be applied to them. However, crystal and molecular structures have very different characteristics. In particular, molecular structures often contain so few atoms that they tend to form low-dimensional local structures, which may hinder the construction of effective frames. Extending our methodology to molecules represents another intriguing future direction for this research.

\section{Conclusions}
This study revisited the challenge of determining effective frames for the SE(3)-invariant modeling of crystal structures. 
We hypothesized that frames should reflect the local dynamic environment around each atom, rather than the static global structure. We herein introduced the concept of \emph{dynamic frames}, which leverage the strengths of interatomic interactions. These frames were integrated into an existing transformer-based network for crystal property prediction~\citep{taniai2024crystalformer}, resulting in the CrystalFramer architecture. Its performance was benchmarked against conventional frame construction methods~\citep{duval23faenet,yan2024comformer} and other state-of-the-art networks~\citep{choudhary21alignn,yan22matformer,lin2023potnet}. Our findings validated the hypothesis,  demonstrating the superior performance of the proposed approach. Although the demonstration focused on crystals, interaction-based dynamic frame construction holds promise for broader applications, including surface modeling, molecular structure modeling, and ML-driven simulations of particles and fluids.

\subsubsection*{Version information}
This arXiv preprint contains the same content as the conference paper (Ito \& Taniai~\etal, 2025) presented at \emph{The Thirteenth International Conference on Learning Representations (ICLR 2025)}, except that some appendices have been moved into the main text to improve readability.

\subsubsection*{Author contributions}
\textbf{YI}, as an intern at OMRON SINIC X, identified the potential utility of the proposed idea for crystals, implemented most of the method in the baseline code, designed and conducted the majority of the experiments, analyzed the results, drafted the initial manuscript, and created the figures.
\textbf{TT} conceived the method, developed and discussed the core idea and philosophy with the co-authors, co-patented the method with YI and RI, explained the baseline code to YI, helped with the coding and experiments, revised the manuscript draft, and led the rebuttal and overall project.
\textbf{RI} provided expertise in materials science and physics simulations, helped review the molecular modeling literature, and advised on the methodology, coding, and writing.
\textbf{YU} co-led materials-related collaborations, provided expertise in machine learning, and advised on the methodology and writing.
\textbf{KO} co-led materials-related collaborations, provided expertise in materials science, and advised on the writing.

\subsubsection*{Acknowledgments}
This work was supported in part by JST-Mirai Program Grant Number JPMJMI21G2 and JST Moonshot R\&D Program Grant Number JPMJMS2236.
TT is supported in part by JSPS KAKENHI Grant Number 24K23911.
RI is supported in part by JSPS KAKENHI Grant Number 24K23910. 
YU is partly supported by JST-Mirai Program Grant Number JPMJMI21G2 and JST Moonshot R\&D Program Grant Number JPMJMS2236.
TT thanks Naoya Chiba for his insightful comment on the weakness of PCA for symmetries, which provided an initial hint for this work.
The authors also thank Yuta Suzuki, Seiji Aota, and the anonymous reviewers for their valuable feedback on the manuscript.

\bibliography{references}

\begin{thebibliography}{54}
\providecommand{\natexlab}[1]{#1}
\providecommand{\url}[1]{\texttt{#1}}
\expandafter\ifx\csname urlstyle\endcsname\relax
  \providecommand{\doi}[1]{doi: #1}\else
  \providecommand{\doi}{doi: \begingroup \urlstyle{rm}\Url}\fi

\bibitem[Batzner et~al.(2022)Batzner, Musaelian, Sun, Geiger, Mailoa, Kornbluth, Molinari, Smidt, and Kozinsky]{Batzner2022nequip}
Simon Batzner, Albert Musaelian, Lixin Sun, Mario Geiger, Jonathan~P. Mailoa, Mordechai Kornbluth, Nicola Molinari, Tess~E. Smidt, and Boris Kozinsky.
\newblock E(3)-equivariant graph neural networks for data-efficient and accurate interatomic potentials.
\newblock \emph{Nature Communications}, 13\penalty0 (1):\penalty0 2453, 2022.
\newblock \doiref{10.1038/s41467-022-29939-5}.

\bibitem[Bihani et~al.(2024)Bihani, Mannan, Pratiush, Du, Chen, Miret, Micoulaut, Smedskjaer, Ranu, and Krishnan]{bihani2024egraff}
Vaibhav Bihani, Sajid Mannan, Utkarsh Pratiush, Tao Du, Zhimin Chen, Santiago Miret, Matthieu Micoulaut, Morten~M. Smedskjaer, Sayan Ranu, and N.~M.~Anoop Krishnan.
\newblock {EGraFFBench}: evaluation of equivariant graph neural network force fields for atomistic simulations.
\newblock \emph{Digital Discovery}, 3:\penalty0 759--768, 2024.
\newblock \doiref{10.1039/D4DD00027G}.

\bibitem[Brandstetter et~al.(2022)Brandstetter, Hesselink, van~der Pol, Bekkers, and Welling]{brandstetter2022segnn}
Johannes Brandstetter, Rob Hesselink, Elise van~der Pol, Erik~J Bekkers, and Max Welling.
\newblock {Geometric and Physical Quantities improve {E(3)} Equivariant Message Passing}.
\newblock In \emph{The Tenth International Conference on Learning Representations (ICLR 2022)}, 2022.
\newblock URL \url{https://openreview.net/forum?id=_xwr8gOBeV1}.

\bibitem[Chanussot et~al.(2021)Chanussot, Das, Goyal, Lavril, Shuaibi, Riviere, Tran, Heras-Domingo, Ho, Hu, Palizhati, Sriram, Wood, Yoon, Parikh, Zitnick, and Ulissi]{chanussot2020oc20}
Lowik Chanussot, Abhishek Das, Siddharth Goyal, Thibaut Lavril, Muhammed Shuaibi, Morgane Riviere, Kevin Tran, Javier Heras-Domingo, Caleb Ho, Weihua Hu, Aini Palizhati, Anuroop Sriram, Brandon Wood, Junwoong Yoon, Devi Parikh, C.~Lawrence Zitnick, and Zachary Ulissi.
\newblock {Open Catalyst 2020 ({OC}20) Dataset and Community Challenges}.
\newblock \emph{ACS Catalysis}, 11\penalty0 (10):\penalty0 6059--6072, 2021.
\newblock \doiref{10.1021/acscatal.0c04525}.

\bibitem[Chen \& Ong(2022)Chen and Ong]{chen22m3gnet}
Chi Chen and Shyue~Ping Ong.
\newblock A universal graph deep learning interatomic potential for the periodic table.
\newblock \emph{Nature Computational Science}, 2\penalty0 (11):\penalty0 718--728, 2022.
\newblock \doiref{10.1038/s43588-022-00349-3}.

\bibitem[Chen et~al.(2019)Chen, Ye, Zuo, Zheng, and Ong]{chen2019megnet}
Chi Chen, Weike Ye, Yunxing Zuo, Chen Zheng, and Shyue~Ping Ong.
\newblock {Graph Networks as a Universal Machine Learning Framework for Molecules and Crystals}.
\newblock \emph{Chemistry of Materials}, 31\penalty0 (9):\penalty0 3564--3572, 2019.
\newblock \doiref{10.1021/acs.chemmater.9b01294}.

\bibitem[Cheng et~al.(2021)Cheng, Zhang, and Dong]{cheng21geocgnn}
Jiucheng Cheng, Chunkai Zhang, and Lifeng Dong.
\newblock A geometric-information-enhanced crystal graph network for predicting properties of materials.
\newblock \emph{Communications Materials}, 2\penalty0 (1):\penalty0 92, 2021.
\newblock \doiref{10.1038/s43246-021-00194-3}.

\bibitem[Chiba et~al.(2023)Chiba, Suzuki, Taniai, Igarashi, Ushiku, Saito, and Ono]{chiba2023nesf}
Naoya Chiba, Yuta Suzuki, Tatsunori Taniai, Ryo Igarashi, Yoshitaka Ushiku, Kotaro Saito, and Kanta Ono.
\newblock Neural structure fields with application to crystal structure autoencoders.
\newblock \emph{Communications Materials}, 4\penalty0 (1):\penalty0 106, 2023.
\newblock \doiref{10.1038/s43246-023-00432-w}.

\bibitem[Choudhary \& DeCost(2021)Choudhary and DeCost]{choudhary21alignn}
Kamal Choudhary and Brian DeCost.
\newblock {Atomistic Line Graph Neural Network for improved materials property predictions}.
\newblock \emph{npj Computational Materials}, 7\penalty0 (1):\penalty0 185, 2021.
\newblock \doiref{10.1038/s41524-021-00650-1}.

\bibitem[Choudhary et~al.(2020)Choudhary, Garrity, Reid, DeCost, Biacchi, Hight~Walker, Trautt, Hattrick-Simpers, Kusne, Centrone, Davydov, Jiang, Pachter, Cheon, Reed, Agrawal, Qian, Sharma, Zhuang, Kalinin, Sumpter, Pilania, Acar, Mandal, Haule, Vanderbilt, Rabe, and Tavazza]{choudhary20jarvis}
Kamal Choudhary, Kevin~F. Garrity, Andrew C.~E. Reid, Brian DeCost, Adam~J. Biacchi, Angela~R. Hight~Walker, Zachary Trautt, Jason Hattrick-Simpers, A.~Gilad Kusne, Andrea Centrone, Albert Davydov, Jie Jiang, Ruth Pachter, Gowoon Cheon, Evan Reed, Ankit Agrawal, Xiaofeng Qian, Vinit Sharma, Houlong Zhuang, Sergei~V. Kalinin, Bobby~G. Sumpter, Ghanshyam Pilania, Pinar Acar, Subhasish Mandal, Kristjan Haule, David Vanderbilt, Karin Rabe, and Francesca Tavazza.
\newblock The joint automated repository for various integrated simulations ({JARVIS}) for data-driven materials design.
\newblock \emph{npj Computational Materials}, 6\penalty0 (1):\penalty0 173, 2020.
\newblock \doiref{10.1038/s41524-020-00440-1}.

\bibitem[Du et~al.(2022)Du, Zhang, Du, Meng, Chen, Zheng, Shao, and Liu]{du2022localframe}
Weitao Du, He~Zhang, Yuanqi Du, Qi~Meng, Wei Chen, Nanning Zheng, Bin Shao, and Tie-Yan Liu.
\newblock {{SE}(3) Equivariant Graph Neural Networks with Complete Local Frames}.
\newblock In \emph{The 39th International Conference on Machine Learning (ICML2022)}, volume 162 of \emph{Proceedings of Machine Learning Research}, pp.\  5583--5608, 2022.
\newblock URL \url{https://proceedings.mlr.press/v162/du22e.html}.

\bibitem[Du et~al.(2023)Du, Du, Wang, Feng, Wang, Ji, Gomes, and Ma]{du2023localframetrans}
Weitao Du, Yuanqi Du, Limei Wang, Dieqiao Feng, Guifeng Wang, Shuiwang Ji, Carla~P Gomes, and Zhi-Ming Ma.
\newblock {A new perspective on building efficient and expressive 3D equivariant graph neural networks}.
\newblock In \emph{Advances in Neural Information Processing Systems (NeurIPS 2023)}, volume~36, pp.\  66647--66674, 2023.
\newblock URL \url{https://papers.nips.cc/paper_files/paper/2023/hash/d212c6c26603c0eb3c9a6b6432386a1f-Abstract-Conference.html}.

\bibitem[Duval et~al.(2024)Duval, Mathis, Joshi, Schmidt, Miret, Malliaros, Cohen, Liò, Bengio, and Bronstein]{duval2024hitchhikersguidegeometricgnns}
Alexandre Duval, Simon~V. Mathis, Chaitanya~K. Joshi, Victor Schmidt, Santiago Miret, Fragkiskos~D. Malliaros, Taco Cohen, Pietro Liò, Yoshua Bengio, and Michael Bronstein.
\newblock {A Hitchhiker's Guide to Geometric GNNs for 3D Atomic Systems}, 2024.
\newblock \arxivref{2312.07511}.

\bibitem[Duval et~al.(2023)Duval, Schmidt, Hern\'{a}ndez-Garc\'{\i}a, Miret, Malliaros, Bengio, and Rolnick]{duval23faenet}
Alexandre~Agm Duval, Victor Schmidt, Alex Hern\'{a}ndez-Garc\'{\i}a, Santiago Miret, Fragkiskos~D. Malliaros, Yoshua Bengio, and David Rolnick.
\newblock {FAENet: Frame Averaging Equivariant GNN for Materials Modeling}.
\newblock In \emph{The 40th International Conference on Machine Learning (ICML 2023)}, volume 202 of \emph{Proceedings of Machine Learning Research}, pp.\  9013--9033, 2023.
\newblock URL \url{https://proceedings.mlr.press/v202/duval23a.html}.

\bibitem[Dym et~al.(2024)Dym, Lawrence, and Siegel]{dym2024continuous}
Nadav Dym, Hannah Lawrence, and Jonathan~W. Siegel.
\newblock {Equivariant Frames and the Impossibility of Continuous Canonicalization}.
\newblock In \emph{The 41st International Conference on Machine Learning (ICML 2024)}, volume 235 of \emph{Proceedings of Machine Learning Research}, pp.\  12228--12267, 2024.
\newblock URL \url{https://proceedings.mlr.press/v235/dym24a.html}.

\bibitem[Fuchs et~al.(2020)Fuchs, Worrall, Fischer, and Welling]{fuchs2020se3transformer}
Fabian Fuchs, Daniel Worrall, Volker Fischer, and Max Welling.
\newblock {SE(3)-Transformers: 3D Roto-Translation Equivariant Attention Networks}.
\newblock In \emph{Advances in Neural Information Processing Systems (NeurIPS 2020)}, volume~33, pp.\  1970--1981, 2020.
\newblock URL \url{https://papers.nips.cc/paper_files/paper/2020/hash/15231a7ce4ba789d13b722cc5c955834-Abstract.html}.

\bibitem[Gerken et~al.(2023)Gerken, Aronsson, Carlsson, Linander, Ohlsson, Petersson, and Persson]{Gerken2023}
Jan~E. Gerken, Jimmy Aronsson, Oscar Carlsson, Hampus Linander, Fredrik Ohlsson, Christoffer Petersson, and Daniel Persson.
\newblock Geometric deep learning and equivariant neural networks.
\newblock \emph{Artificial Intelligence Review}, 56\penalty0 (12):\penalty0 14605--14662, 2023.
\newblock \doiref{10.1007/s10462-023-10502-7}.

\bibitem[Han et~al.(2024)Han, Cen, Wu, Li, Kong, Jiao, Yu, Xu, Wu, Wang, Xu, Wei, Liu, Rong, and Huang]{han2024surveygeometricgraphneural}
Jiaqi Han, Jiacheng Cen, Liming Wu, Zongzhao Li, Xiangzhe Kong, Rui Jiao, Ziyang Yu, Tingyang Xu, Fandi Wu, Zihe Wang, Hongteng Xu, Zhewei Wei, Yang Liu, Yu~Rong, and Wenbing Huang.
\newblock {A Survey of Geometric Graph Neural Networks: Data Structures, Models and Applications}, 2024.
\newblock \arxivref{2403.00485}.

\bibitem[Huang et~al.(2020)Huang, Perez, Ba, and Volkovs]{huang20improving}
Xiao~Shi Huang, Felipe Perez, Jimmy Ba, and Maksims Volkovs.
\newblock {Improving Transformer Optimization Through Better Initialization}.
\newblock In \emph{The 37th International Conference on Machine Learning (ICML 2020)}, volume 119 of \emph{Proceedings of Machine Learning Research}, pp.\  4475--4483, 2020.
\newblock URL \url{https://proceedings.mlr.press/v119/huang20f.html}.

\bibitem[Izmailov et~al.(2018)Izmailov, Podoprikhin, Garipov, Vetrov, and Wilson]{izmailov2018averaging}
Pavel Izmailov, Dmitrii Podoprikhin, Timur Garipov, Dmitry~P. Vetrov, and Andrew~Gordon Wilson.
\newblock {Averaging Weights Leads to Wider Optima and Better Generalization}.
\newblock In \emph{The 34th Conference on Uncertainty in Artificial Intelligence (UAI 2018)}, pp.\  876--885, 2018.
\newblock URL \url{http://auai.org/uai2018/proceedings/papers/313.pdf}.

\bibitem[Jain et~al.(2013)Jain, Ong, Hautier, Chen, Richards, Dacek, Cholia, Gunter, Skinner, Ceder, and Persson]{jain2013materialsproject}
Anubhav Jain, Shyue~Ping Ong, Geoffroy Hautier, Wei Chen, William~Davidson Richards, Stephen Dacek, Shreyas Cholia, Dan Gunter, David Skinner, Gerbrand Ceder, and Kristin~A. Persson.
\newblock {The Materials Project: A materials genome approach to accelerating materials innovation}.
\newblock \emph{APL Materials}, 1\penalty0 (1):\penalty0 011002, 2013.
\newblock \doiref{10.1063/1.4812323}.

\bibitem[Jiao et~al.(2023)Jiao, Huang, Lin, Han, Chen, Lu, and Liu]{jiao2023crystalpredict}
Rui Jiao, Wenbing Huang, Peijia Lin, Jiaqi Han, Pin Chen, Yutong Lu, and Yang Liu.
\newblock {Crystal Structure Prediction by Joint Equivariant Diffusion}.
\newblock In \emph{Advances in Neural Information Processing Systems (NeurIPS 2023)}, volume~36, pp.\  17464--17497, 2023.
\newblock URL \url{https://proceedings.neurips.cc/paper_files/paper/2023/hash/38b787fc530d0b31825827e2cc306656-Abstract-Conference.html}.

\bibitem[Kingma \& Ba(2015)Kingma and Ba]{kingma14adam}
Diederik~P. Kingma and Jimmy Ba.
\newblock {Adam: {A} Method for Stochastic Optimization}.
\newblock In \emph{The Third International Conference on Learning Representations (ICLR 2015)}, 2015.
\newblock \arxivref{1412.6980}.

\bibitem[Kirklin et~al.(2015)Kirklin, Saal, Meredig, Thompson, Doak, Aykol, R{\"u}hl, and Wolverton]{Kirklin2015oqmd}
Scott Kirklin, James~E. Saal, Bryce Meredig, Alex Thompson, Jeff~W. Doak, Muratahan Aykol, Stephan R{\"u}hl, and Chris Wolverton.
\newblock {The Open Quantum Materials Database (OQMD): assessing the accuracy of DFT formation energies}.
\newblock \emph{npj Computational Materials}, 1\penalty0 (1):\penalty0 15010, 2015.
\newblock \doiref{10.1038/npjcompumats.2015.10}.

\bibitem[Liao \& Smidt(2023)Liao and Smidt]{liao2023equiformer}
Yi-Lun Liao and Tess Smidt.
\newblock {Equiformer: Equivariant Graph Attention Transformer for 3D Atomistic Graphs}.
\newblock In \emph{The Eleventh International Conference on Learning Representations (ICLR 2023)}, 2023.
\newblock URL \url{https://openreview.net/forum?id=KwmPfARgOTD}.

\bibitem[Liao et~al.(2024)Liao, Wood, Das, and Smidt]{liao2024equiformerv2}
Yi-Lun Liao, Brandon~M Wood, Abhishek Das, and Tess Smidt.
\newblock {EquiformerV2: Improved Equivariant Transformer for Scaling to Higher-Degree Representations}.
\newblock In \emph{The Twelfth International Conference on Learning Representations (ICLR 2024)}, 2024.
\newblock URL \url{https://openreview.net/forum?id=mCOBKZmrzD}.

\bibitem[Lin et~al.(2022)Lin, Wang, Liu, and Qiu]{lin2021surveytransformers}
Tianyang Lin, Yuxin Wang, Xiangyang Liu, and Xipeng Qiu.
\newblock A survey of transformers.
\newblock \emph{AI Open}, 3:\penalty0 111--132, 2022.
\newblock \doiref{10.1016/j.aiopen.2022.10.001}.

\bibitem[Lin et~al.(2023)Lin, Yan, Luo, Liu, Qian, and Ji]{lin2023potnet}
Yuchao Lin, Keqiang Yan, Youzhi Luo, Yi~Liu, Xiaoning Qian, and Shuiwang Ji.
\newblock {Efficient Approximations of Complete Interatomic Potentials for Crystal Property Prediction}.
\newblock In \emph{The 40th International Conference on Machine Learning (ICML 2023)}, volume 202 of \emph{Proceedings of Machine Learning Research}, pp.\  21260--21287, 2023.
\newblock URL \url{https://proceedings.mlr.press/v202/lin23m.html}.

\bibitem[Lin et~al.(2024)Lin, Helwig, Gui, and Ji]{lin2024minimalfa}
Yuchao Lin, Jacob Helwig, Shurui Gui, and Shuiwang Ji.
\newblock {Equivariance via Minimal Frame Averaging for More Symmetries and Efficiency}.
\newblock In \emph{The 41st International Conference on Machine Learning (ICML 2024)}, volume 235 of \emph{Proceedings of Machine Learning Research}, pp.\  30042--30079, 2024.
\newblock URL \url{https://proceedings.mlr.press/v235/lin24i.html}.

\bibitem[Liu et~al.(2024)Liu, Yao, Zhan, Ma, Pan, and Hu]{liu2024gradformer}
Chuang Liu, Zelin Yao, Yibing Zhan, Xueqi Ma, Shirui Pan, and Wenbin Hu.
\newblock {Gradformer: Graph Transformer with Exponential Decay}.
\newblock In \emph{The 33rd International Joint Conference on Artificial Intelligence (IJCAI 2024)}, pp.\  2171--2179, 2024.
\newblock Main Track. \doiref{10.24963/ijcai.2024/240}.

\bibitem[Loshchilov \& Hutter(2019)Loshchilov and Hutter]{loshchilov19adamw}
Ilya Loshchilov and Frank Hutter.
\newblock {Decoupled Weight Decay Regularization}.
\newblock In \emph{The Seventh International Conference on Learning Representations (ICLR 2019)}, 2019.
\newblock URL \url{https://openreview.net/forum?id=Bkg6RiCqY7}.

\bibitem[Louis et~al.(2020)Louis, Zhao, Nasiri, Wang, Song, Liu, and Hu]{louis20gatgnn}
Steph-Yves Louis, Yong Zhao, Alireza Nasiri, Xiran Wang, Yuqi Song, Fei Liu, and Jianjun Hu.
\newblock Graph convolutional neural networks with global attention for improved materials property prediction.
\newblock \emph{Physical Chemistry Chemical Physics}, 22:\penalty0 18141--18148, 2020.
\newblock \doiref{10.1039/D0CP01474E}.

\bibitem[Niggli(1928)]{niggli}
Paul Niggli.
\newblock \emph{Handbuch der Experimentalphysik}.
\newblock akademische Verlagsgesellschaft, 1928.

\bibitem[Park \& Wolverton(2020)Park and Wolverton]{park20icgcnn}
Cheol~Woo Park and Chris Wolverton.
\newblock Developing an improved crystal graph convolutional neural network framework for accelerated materials discovery.
\newblock \emph{Physical Review Materials}, 4:\penalty0 063801, 2020.
\newblock \doiref{10.1103/PhysRevMaterials.4.063801}.

\bibitem[Pozdnyakov \& Ceriotti(2023)Pozdnyakov and Ceriotti]{pozdnyakov2023ensemble}
Sergey Pozdnyakov and Michele Ceriotti.
\newblock Smooth, exact rotational symmetrization for deep learning on point clouds.
\newblock In \emph{Advances in Neural Information Processing Systems (NeurIPS 2023)}, volume~36, pp.\  79469--79501, 2023.
\newblock URL \url{https://proceedings.neurips.cc/paper_files/paper/2023/hash/fb4a7e3522363907b26a86cc5be627ac-Abstract-Conference.html}.

\bibitem[Pozdnyakov \& Ceriotti(2022)Pozdnyakov and Ceriotti]{pozdnyakov22incomplete}
Sergey~N Pozdnyakov and Michele Ceriotti.
\newblock Incompleteness of graph neural networks for points clouds in three dimensions.
\newblock \emph{Machine Learning: Science and Technology}, 3\penalty0 (4):\penalty0 045020, 2022.
\newblock \doiref{10.1088/2632-2153/aca1f8}.

\bibitem[Puny et~al.(2022)Puny, Atzmon, Smith, Misra, Grover, Ben-Hamu, and Lipman]{puny2022frameaveraging}
Omri Puny, Matan Atzmon, Edward~J. Smith, Ishan Misra, Aditya Grover, Heli Ben-Hamu, and Yaron Lipman.
\newblock {Frame Averaging for Invariant and Equivariant Network Design}.
\newblock In \emph{The Tenth International Conference on Learning Representations (ICLR 2022)}, 2022.
\newblock URL \url{https://openreview.net/forum?id=zIUyj55nXR}.

\bibitem[Qi et~al.(2017)Qi, Su, Mo, and Guibas]{qi17pointnet}
Charles~Ruizhongtai Qi, Hao Su, Kaichun Mo, and Leonidas~J. Guibas.
\newblock {PointNet: Deep Learning on Point Sets for 3D Classification and Segmentation}.
\newblock In \emph{2017 {IEEE} Conference on Computer Vision and Pattern Recognition (CVPR 2017)}, pp.\  77--85, 2017.
\newblock \doiref{10.1109/CVPR.2017.16}. URL \url{https://openaccess.thecvf.com/content_cvpr_2017/html/Qi_PointNet_Deep_Learning_CVPR_2017_paper.html}.

\bibitem[Santoro \& Mighell(1970)Santoro and Mighell]{niggli_algorithm}
A.~Santoro and A.~D. Mighell.
\newblock Determination of reduced cells.
\newblock \emph{Acta Crystallographica Section A}, 26\penalty0 (1):\penalty0 124--127, 1970.
\newblock \doiref{10.1107/S0567739470000177}.

\bibitem[Sch\"{u}tt et~al.(2018)Sch\"{u}tt, Sauceda, Kindermans, Tkatchenko, and Müller]{schutt18schnet}
K.~T. Sch\"{u}tt, H.~E. Sauceda, P.-J. Kindermans, A.~Tkatchenko, and K.-R. Müller.
\newblock {SchNet -- A deep learning architecture for molecules and materials}.
\newblock \emph{The Journal of Chemical Physics}, 148\penalty0 (24):\penalty0 241722, 2018.
\newblock \doiref{10.1063/1.5019779}.

\bibitem[Shaw et~al.(2018)Shaw, Uszkoreit, and Vaswani]{shaw18relative}
Peter Shaw, Jakob Uszkoreit, and Ashish Vaswani.
\newblock {Self-Attention with Relative Position Representations}.
\newblock In \emph{The 2018 Conference of the North {A}merican Chapter of the Association for Computational Linguistics: Human Language Technologies (NAACL-HLT 2018)}, volume 2 (Short Papers), pp.\  464--468, 2018.
\newblock \doiref{10.18653/v1/N18-2074}.

\bibitem[Shi et~al.(2023)Shi, Zheng, Ke, Shen, You, He, Luo, Liu, He, and Liu]{shi2023graphormer3d}
Yu~Shi, Shuxin Zheng, Guolin Ke, Yifei Shen, Jiacheng You, Jiyan He, Shengjie Luo, Chang Liu, Di~He, and Tie-Yan Liu.
\newblock {Benchmarking Graphormer on Large-Scale Molecular Modeling Datasets}, 2023.
\newblock \arxivref{2203.04810}.

\bibitem[Suzuki et~al.(2022)Suzuki, Taniai, Saito, Ushiku, and Ono]{Suzuki2022dml}
Yuta Suzuki, Tatsunori Taniai, Kotaro Saito, Yoshitaka Ushiku, and Kanta Ono.
\newblock Self-supervised learning of materials concepts from crystal structures via deep neural networks.
\newblock \emph{Machine Learning: Science and Technology}, 3\penalty0 (4):\penalty0 045034, 2022.
\newblock \doiref{10.1088/2632-2153/aca23d}.

\bibitem[Suzuki et~al.(2025)Suzuki, Taniai, Igarashi, Saito, Chiba, Ushiku, and Ono]{suzuki2025contrastive}
Yuta Suzuki, Tatsunori Taniai, Ryo Igarashi, Kotaro Saito, Naoya Chiba, Yoshitaka Ushiku, and Kanta Ono.
\newblock {Contrastive Language-Structure Pre-training Driven by Materials Science Literature}, 2025.
\newblock \arxivref{2501.12919}.

\bibitem[Taniai et~al.(2024)Taniai, Igarashi, Suzuki, Chiba, Saito, Ushiku, and Ono]{taniai2024crystalformer}
Tatsunori Taniai, Ryo Igarashi, Yuta Suzuki, Naoya Chiba, Kotaro Saito, Yoshitaka Ushiku, and Kanta Ono.
\newblock {Crystalformer: Infinitely Connected Attention for Periodic Structure Encoding}.
\newblock In \emph{The Twelfth International Conference on Learning Representations (ICLR 2024)}, 2024.
\newblock URL \url{https://openreview.net/forum?id=fxQiecl9HB}.

\bibitem[Thomas et~al.(2018)Thomas, Smidt, Kearnes, Yang, Li, Kohlhoff, and Riley]{thomas2018tfn}
Nathaniel Thomas, Tess Smidt, Steven Kearnes, Lusann Yang, Li~Li, Kai Kohlhoff, and Patrick Riley.
\newblock {Tensor field networks: Rotation- and translation-equivariant neural networks for 3D point clouds}, 2018.
\newblock \arxivref{1802.08219}.

\bibitem[Tran et~al.(2023)Tran, Lan, Shuaibi, Wood, Goyal, Das, Heras-Domingo, Kolluru, Rizvi, Shoghi, Sriram, Therrien, Abed, Voznyy, Sargent, Ulissi, and Zitnick]{tran2022oc22}
Richard Tran, Janice Lan, Muhammed Shuaibi, Brandon~M. Wood, Siddharth Goyal, Abhishek Das, Javier Heras-Domingo, Adeesh Kolluru, Ammar Rizvi, Nima Shoghi, Anuroop Sriram, F{\'e}lix Therrien, Jehad Abed, Oleksandr Voznyy, Edward~H. Sargent, Zachary Ulissi, and C.~Lawrence Zitnick.
\newblock {The Open Catalyst 2022 (OC22) Dataset and Challenges for Oxide Electrocatalysts}.
\newblock \emph{ACS Catalysis}, 13\penalty0 (5):\penalty0 3066--3084, 2023.
\newblock \doiref{10.1021/acscatal.2c05426}.

\bibitem[Vaswani et~al.(2017)Vaswani, Shazeer, Parmar, Uszkoreit, Jones, Gomez, Kaiser, and Polosukhin]{vaswani17transformer}
Ashish Vaswani, Noam Shazeer, Niki Parmar, Jakob Uszkoreit, Llion Jones, Aidan~N Gomez, {\L}ukasz Kaiser, and Illia Polosukhin.
\newblock {Attention is All you Need}.
\newblock In \emph{Advances in Neural Information Processing Systems (NIPS 2017)}, volume~30, pp.\  6000--6010, 2017.
\newblock URL \url{https://proceedings.neurips.cc/paper_files/paper/2017/hash/3f5ee243547dee91fbd053c1c4a845aa-Abstract.html}.

\bibitem[Wang et~al.(2023)Wang, Li, Wang, Shao, Zheng, and Liu]{wang2023geoformer}
Yusong Wang, Shaoning Li, Tong Wang, Bin Shao, Nanning Zheng, and Tie-Yan Liu.
\newblock {Geometric Transformer with Interatomic Positional Encoding}.
\newblock In \emph{Advances in Neural Information Processing Systems (NeurIPS 2023)}, volume~36, pp.\  55981--55994, 2023.
\newblock URL \url{https://proceedings.neurips.cc/paper_files/paper/2023/hash/aee2f03ecb2b2c1ea55a43946b651cfd-Abstract-Conference.html}.

\bibitem[Xie \& Grossman(2018)Xie and Grossman]{xie18cgcnn}
Tian Xie and Jeffrey~C. Grossman.
\newblock {Crystal Graph Convolutional Neural Networks for an Accurate and Interpretable Prediction of Material Properties}.
\newblock \emph{Physical Review Letters}, 120:\penalty0 145301, 2018.
\newblock \doiref{10.1103/PhysRevLett.120.145301}.

\bibitem[Yan et~al.(2022)Yan, Liu, Lin, and Ji]{yan22matformer}
Keqiang Yan, Yi~Liu, Yuchao Lin, and Shuiwang Ji.
\newblock {Periodic Graph Transformers for Crystal Material Property Prediction}.
\newblock In \emph{Advances in Neural Information Processing Systems (NeurIPS 2022)}, volume~35, pp.\  15066--15080, 2022.
\newblock URL \url{https://proceedings.neurips.cc/paper_files/paper/2022/hash/6145c70a4a4bf353a31ac5496a72a72d-Abstract-Conference.html}.

\bibitem[Yan et~al.(2024)Yan, Fu, Qian, Qian, and Ji]{yan2024comformer}
Keqiang Yan, Cong Fu, Xiaofeng Qian, Xiaoning Qian, and Shuiwang Ji.
\newblock {Complete and Efficient Graph Transformers for Crystal Material Property Prediction}.
\newblock In \emph{The Twelfth International Conference on Learning Representations (ICLR 2024)}, 2024.
\newblock URL \url{https://openreview.net/forum?id=BnQY9XiRAS}.

\bibitem[Ying et~al.(2021)Ying, Cai, Luo, Zheng, Ke, He, Shen, and Liu]{ying21graphormer}
Chengxuan Ying, Tianle Cai, Shengjie Luo, Shuxin Zheng, Guolin Ke, Di~He, Yanming Shen, and Tie-Yan Liu.
\newblock {Do Transformers Really Perform Badly for Graph Representation?}
\newblock In \emph{Advances in Neural Information Processing Systems (NeurIPS 2021)}, volume~34, pp.\  28877--28888, 2021.
\newblock URL \url{https://proceedings.neurips.cc/paper_files/paper/2021/hash/f1c1592588411002af340cbaedd6fc33-Abstract.html}.

\bibitem[Zaheer et~al.(2017)Zaheer, Kottur, Ravanbakhsh, Poczos, Salakhutdinov, and Smola]{zaheer17deepsets}
Manzil Zaheer, Satwik Kottur, Siamak Ravanbakhsh, Barnabas Poczos, Russ~R Salakhutdinov, and Alexander~J Smola.
\newblock {Deep Sets}.
\newblock In \emph{Advances in Neural Information Processing Systems (NIPS 2017)}, volume~30, pp.\  3394--3404, 2017.
\newblock URL \url{https://proceedings.neurips.cc/paper_files/paper/2017/hash/f22e4747da1aa27e363d86d40ff442fe-Abstract.html}.

\end{thebibliography}
\bibliographystyle{iclr2025_conference}
\appendix
\renewcommand{\theequation}{A\arabic{equation}}
\setcounter{equation}{0}
\renewcommand{\thetable}{A\arabic{table}}
\setcounter{table}{0}
\renewcommand{\thefigure}{A\arabic{figure}}
\setcounter{figure}{0}
\renewcommand{\theHfigure}{A\arabic{figure}}
\renewcommand{\theHtable}{A\arabic{table}}
\renewcommand{\theHequation}{A\arabic{equation}}

\section{Limitations of unit-cell-based crystal representations}
\label{appendix:primitive_cell}
The conventional PCA frames explained in Sec.~\ref{sec:existing_frames} implicitly assume a unique lattice representation such as the Niggli reduced cell~\citep{niggli}. Similarly, the lattice frames assume a primitive cell and convert it to a cell similar to the reduced one. 
Otherwise, these frames are affected by the arbitrariness of unit cell representations, such as supercells and conventional cells.

Traditionally, primitive cells and conventional cells are used to represent periodic structures. Primitive cells are defined as the smallest repeating units of a lattice, having the minimum volume and containing only a single lattice point within each cell. By following a mathematical procedure on primitive cells, their unique representations called reduced unit cells can be obtained~\citep{niggli_algorithm}.

On the other hand, conventional cells are defined as unit cells that are not necessarily primitive but are designed to exhibit symmetry in an easily understandable way. The notion of conventional cells is often illustrated by the face-centered cubic lattice and the body-centered cubic lattice.

\begin{figure}[h]
\centering
\includegraphics[width=0.4\textwidth]{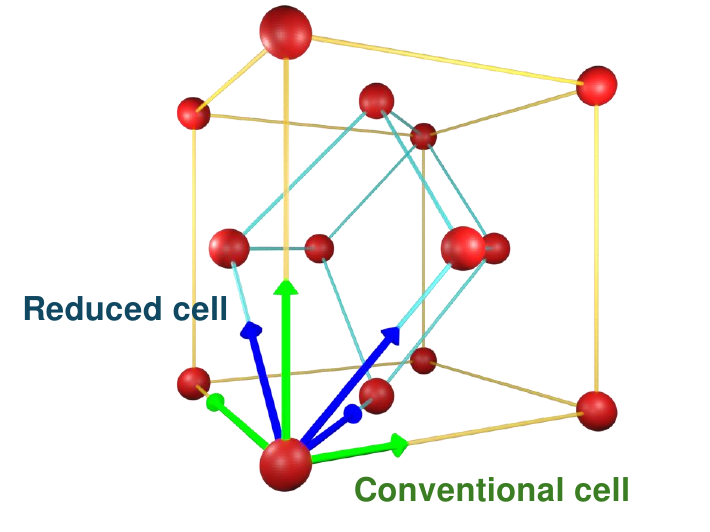}
\caption{\textbf{Conventional cell (green) and Niggli reduced cell (blue) for a face-centered cube.}}
\label{fig:appendix_fcc}
\end{figure}

Figure~\ref{fig:appendix_fcc} compares a conventional cell and the Niggli reduced cell of a face-centered cubic structure. Examining the conventional unit cell easily reveals that it represents a cubic lattice, with atoms located at each corner and at each face center. However, this fact is obscured in the reduced cell. Therefore, reduced cells can be said to sacrifice the interpretability of physically important information, such as symmetry, in order to uniquely represent periodic structures. 

\section{Comparison to existing local frames for molecules}
\label{appendix:localframe}
In the molecular modeling literature, several local frames have been proposed~\citep{du2022localframe,du2023localframetrans,pozdnyakov2023ensemble}.
The concept of our dynamic frames, being both dynamic and local, is distinct from these frames for molecules, which are local but static. Below we discuss this perspective in more detail.

We first clarify the terminology regarding `dynamic' and `static' in this context. We use `dynamic' to describe behavior that is influenced by the model's internal states estimated for a given structure. For instance, interatomic interactions modeled within a GNN reflect these internal states and evolve dynamically layer by layer. Dynamic frames are designed to align with these interatomic interactions. While the molecular modeling literature often uses `dynamic' to describe temporally evolving structures, our work does not assume such temporal dynamics. Similarly, we use `static' to describe behavior that is unaffected by the model’s internal states.

\textbf{\cite{du2022localframe} propose static edge-wise frames.} These edge-wise frames, denoted as $F_{ij} = [\bm{e}_1, \bm{e}_2, \bm{e}_3]^T$ using our notation, are $3\times3$ orthogonal matrices defined individually for each edge $(i, j)$. From Eq.~2 in their paper, the axes of $F_{ij}$ are defined as $\bm{e}_1 = \text{unit}(\bm{p}_i-\bm{p}_j)$, $\bm{e}_2 = \text{unit}(\bm{p}_i\times \bm{p}_j)$, and $\bm{e}_3 = \bm{e}_1 \times \bm{e}_2$, where $\text{unit}(\bm{x}) = \bm{x}/\|\bm{x}\|$ is $L_2$ normalization. Here, the centroid of the structure is pre-shifted to the origin, as $\bm{p} \gets \bm{p} - \bar{\bm{p}}$ using $\bar{\bm{p}} = \frac{1}{N}\sum_{i} \bm{p}_i$. Thus, these frames are translation invariant, even though $\bm{e}_2$ appears to depend on absolute positions. However, performing such a global centroid shift for crystals is not straightforward due to their infinite periodicity, unless a specific unit cell description is utilized.

\textbf{\cite{du2023localframetrans} propose frame-based equivariant message passing using static edge-wise and node-wise frames.} These edge-wise frames $F_{ij}$ are identical to those used in their earlier work~\citep{du2022localframe}~(see above). Their node-wise frames $F_i$ are defined similarly to $F_{ij}$, but with $\bm{p}_j$ replaced by the cluster centroid around $i$: $\bar{\bm{p}}_i = \frac{1}{|N(i)|}\sum_{j \in N(i)} \bm{p}_j$. Thus, the axes of $F_i$ are provided as $\bm{e}_1 = \text{unit}(\bm{p}_i-\bar{\bm{p}}_i)$, $\bm{e}_2 = \text{unit}(\bm{p}_i\times \bar{\bm{p}}_i)$, and $\bm{e}_3 = \bm{e}_1 \times \bm{e}_2$. (See Eqs.~13 and 14 in their paper for the definitions.) To ensure translation invariance, these node-wise frames also rely on global centroid normalization. Moreover, when applied to crystal structures. their highly symmetric nature will often cause $\bar{\bm{p}}_i \simeq \bm{p}_i$, resulting in unstable frame construction.

\textbf{\cite{pozdnyakov2023ensemble} propose ensemble of many 3-body interactions} called the equivariant coordinate-system ensemble. For each target atom $i$, they construct many triplets of atoms $(i, j, j')$ using pairs of neighbors $(j, j')$ and then construct a local frame for each triplet as $F_{ijj'}$. Although these triplet-wise frames are local, they do not reflect dynamic internal states of the model. Also, modeling 3-body interactions is computationally expensive.

Overall, these methods all employ specific types of static local frames, such as node-wise~\citep{du2023localframetrans}, edge-wise~\citep{du2022localframe,du2023localframetrans}, or triplet-wise~\citep{pozdnyakov2023ensemble} frames.
None of them leverage the model's internal states for frame construction.

In Sec.~\ref{sec:property_prediction}, we further compare the proposed method using dynamic frames with its static counterpart variant, which is based on static local frames. The results in Tables~\ref{table:results_jarvis} and \ref{table:results_mp} demonstrate the superior performance of the proposed dynamic frames, 
highlighting the conceptual difference between these two families of frames.

\section{Dataset specifications}
\label{appendix:dataset}
We use the following three sources of materials data for evaluations. They are all publicly available through a Python package (jarvis-tools) created by \cite{choudhary20jarvis}.

\textbf{The JARVIS-DFT 3D 2021} is a collection of 55,723 materials provided by \cite{choudhary20jarvis} and is accessible as \texttt{dft\_3d\_2021} via {jarvis-tools} (or as \texttt{dft\_3d} in older versions). These materials are annotated with various simulated properties using two DFT calculation methods, OptB88vdW (OPT) and TBmBJ (MBJ). Following recent studies~\citep{yan22matformer,yan2024comformer,lin2023potnet,taniai2024crystalformer}, we use formation energy (\texttt{formation\_energy\_peratom}), total energy (\texttt{optb88vdw\_total\_energy}), bandgap (\texttt{optb88vdw\_bandgap} and \texttt{mbj\_bandgap}), and energy above hull or E hull (\texttt{ehull}) as regression targets.

\textbf{The Materials Project (MP) database}~\citep{jain2013materialsproject} is an online public materials database providing various synthetic materials and their DFT-calculated properties. We specifically use its snapshot collected by \cite{chen2019megnet}, which contains 69,239 materials and is accessible as \texttt{megnet} via jarvis-tools. Following recent studies~\citep{yan22matformer,yan2024comformer,lin2023potnet,taniai2024crystalformer}, we use formation energy (\texttt{e\_form}), bandgap (\texttt{gap pbe}), bulk modulus (\texttt{bulk modulus}), and shear modulus (\texttt{shear modulus}) as regression targets. For bulk and shear modulus, we use the data splits provided by \cite{yan22matformer}.

\textbf{The Open Quantum Materials Database (OQMD)} is another online public materials database by \cite{Kirklin2015oqmd}. We specifically use its snapshot provided as \texttt{oqmd\_3d\_no\_cfid} in jarvis-tools, which contains 817,636 materials with three DFT-calculated properties: formation energy (\texttt{\_oqmd\_delta\_e}), bandgap (\texttt{\_oqmd\_band\_gap}), and energy above hull (\texttt{\_oqmd\_stability}). We use these properties as regression targets. We release our data splits along with our code.

\section{Training settings}
\label{appendix:detail_train_param}
Table~\ref{appendix:table:detailed_param} summarizes the training settings for the JARVIS, MP, and OQMD datasets.

Specifically, for the JARVIS dataset, we optimize the mean absolute loss function using the Adam optimizer~\citep{kingma14adam} with $(\beta_1, \beta_2) = (0.9, 0.98)$ and weight decay of $10^{-5}$~\citep{loshchilov19adamw}. 
We employ the warm-up-free inverse square root scheduling~\citep{huang20improving} for the learning rate, with the initial learning rate of $5.0\times 10^{-4}$ and decay factor of $ \sqrt{4000/(4000+t)}$ according to the total train steps $t$.
The model weights are initialized through the strategy for the normalization-free transformer architecture by \cite{huang20improving}, which improves the training stability. 
The training is iterated for a total of 2000 epochs with a batch size of 256 materials. 
Stochastic weight averaging (SWA)~\citep{izmailov2018averaging} is adopted for model selection for testing and validation. Except for the increased number of epochs, we use the same settings with the baseline Crystalformer model~\citep{taniai2024crystalformer} to evaluate the pure effects of introducing the frames.

For the OQMD dataset, which was not used by the baseline method, we use similar settings with a larger batch size of 1024 materials and fewer epochs of 200.

\begin{table}[h]
\centering
\caption{\textbf{Detailed training settings.}}
  \vskip 0.5mm
\scalebox{0.8}{
\begin{tabular}{ll}
\toprule[1.2pt]
Hyperparameters & Settings (JARVIS, MP, OQMD)
 \\
\midrule[1.2pt]
Loss function & Mean absolute error \\
Optimizer & AdamW with $(\beta_1, \beta_2) = (0.9, 0.98)$ \\
Weight decay & $10^{-5}$ \\
Gradient norm clipping & 1.0 \\
Initial learning rate $\alpha$ & $5.0 \times 10 ^{-4}$ \\
Learning rate scheduling per step & $\alpha \sqrt{4000/(4000+ t)}$\\
Warm-up steps & $0$ (no warm-up) \\
Batch size & $256$, $128$, $1024$ \\
Number of epochs & $2000$, $800$, $200$ \\
Dropout rate & $0.0$ \\
SWA epochs & $50$, $50$, $20$ \\
\bottomrule[1.2pt]
\end{tabular}
}
\label{appendix:table:detailed_param}
\end{table}

\section{Evolution of dynamic frames during training}
\label{appendix:evolution}
\begin{figure}[t]
\centering
\includegraphics[width=0.75\textwidth]{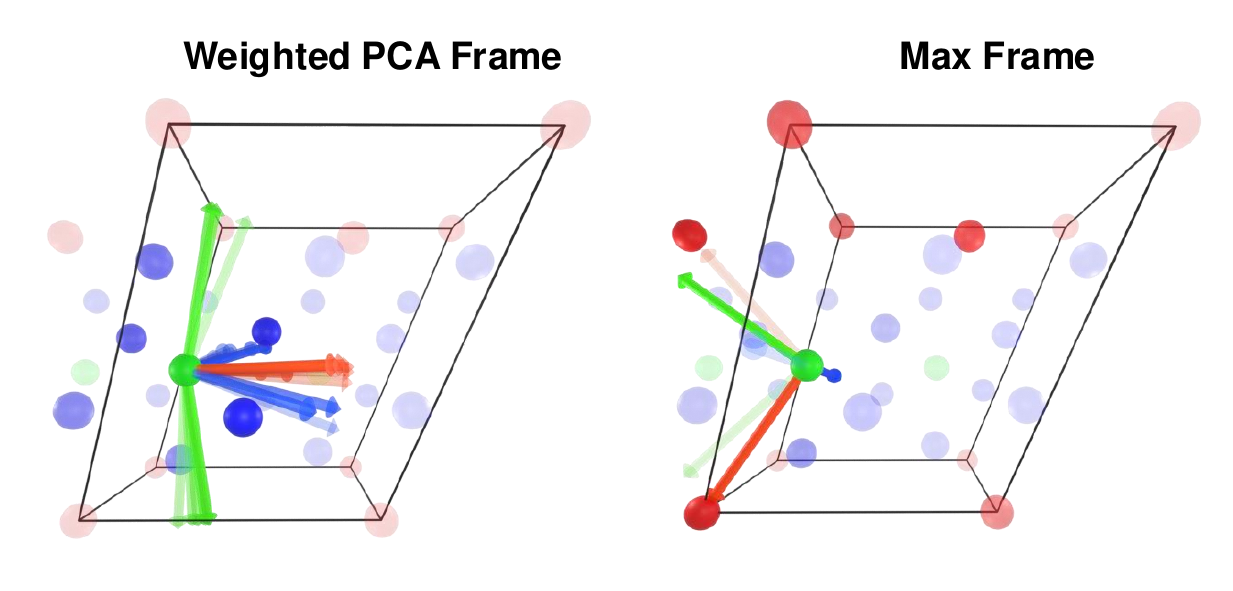}
\caption{\textbf{Evolution of dynamic frames during training.} We visualize the weighted PCA frames and max frames using model checkpoints taken every 200 epochs, starting from epoch 100 until 2000. Frames from earlier checkpoints are overlaid with higher transparency. Notably, the max frames stabilize more quickly than the weighted PCA frames.}
\label{fig:appendix_evolution}
\end{figure}

We examined how dynamic frames evolve throughout the training process, by visualizing frames using model checkpoints taken at 200-epoch intervals.
Figure~\ref{fig:appendix_evolution} compares the evolution of the weighted PCA frames and max frames for the same material as Fig.~\ref{fig:frame_comparison}a. We observed that the weighted PCA frames fluctuated throughout training, whereas the max frames stabilized quickly during the early stages. As frame fluctuations can introduce noise, the early stabilization of the max frames may explain their superior performance compared to the weighted PCA frames.

\section{Scalability for large structures and supercells}
\label{appendix:scalability}
Since the proposed CrystalFramer is based on a self-attention mechanism, its computational complexity is $O(Nk)$, where $N$ is the number of atoms in the unit cell and $k$ is the number of neighbors per unit-cell atom. 
In the infinitely connected attention of Crystalformer~\citep{taniai2024crystalformer} defined in Eq.~\ref{eq:attention}, neighbors $j(\bm{n})$ are adaptively determined for each atom $i$ in each layer. The current implementation computes neighbors by periodically repeating the unit cell within a finite range. Consequently, $k$ becomes a multiple of $N$, resulting in an overall computational complexity of $O(N^2)$. 

In practice, the training of CrystalFramer has successfully scaled to relatively large structures in the MP dataset, which features an average of 30 atoms per unit cell and a maximum of 296 atoms. For inference, the method can handle even larger structures than during training, as it requires significantly less memory and supports per-material (non-batched) processing.

Scalability for larger structures becomes crucial especially when processing supercells. Supercells are often utilized when structures deviate from perfect periodicity, such as in the presence of impurities, defects, or surfaces. We consider the following two potential approaches to improve efficiency with large supercells.

\textbf{Mixed atom embedding.} Structures with impurities or defects are often represented using site occupancy, which indicates the probabilities of different elements occupying an atomic site. Instead of modeling such structures with supercells, we can efficiently represent the site occupancy by mixing atomic embedding vectors. In this case, each $a_i$ represents a probability distribution over elements rather than a single element. The corresponding atomic state can then be initialized as a linear blend of atom embeddings: $\bm{x}_i \gets \sum_{\text{element}} a_i(\text{element}) \text{AtomEmbedding}(\text{element})$. This approach can keep the structure size small without using a supercell, thereby maintaining overall efficiency.

\textbf{Distance-based neighbor search.} When unit cells are large, the current cell-based neighbor identification method will produce redundant neighbors, forcing $k \ge N$. By employing a more compact set of neighbors through  nearest neighbor search, the complexity is reduced from $O(N^2)$ to $O(Nk)$, improving efficiency for larger structures.

Since structures with imperfect periodicity are common in realistic scenarios, developing scalable models for these structures is an important direction for future research.

\section{Analysis of model's continuity}
\label{appendix:continuity}
\cite{dym2024continuous} pointed out that frame-based models generally exhibit discontinuous characteristics, which are also inherent in our approach. To empirically assess the degree of this discontinuity in our trained models, we analyze the variations in their outputs for a given crystal structure under perturbations.

\begin{figure}[t]
\vskip 2\baselineskip
\centering
\includegraphics[width=1.0\textwidth]{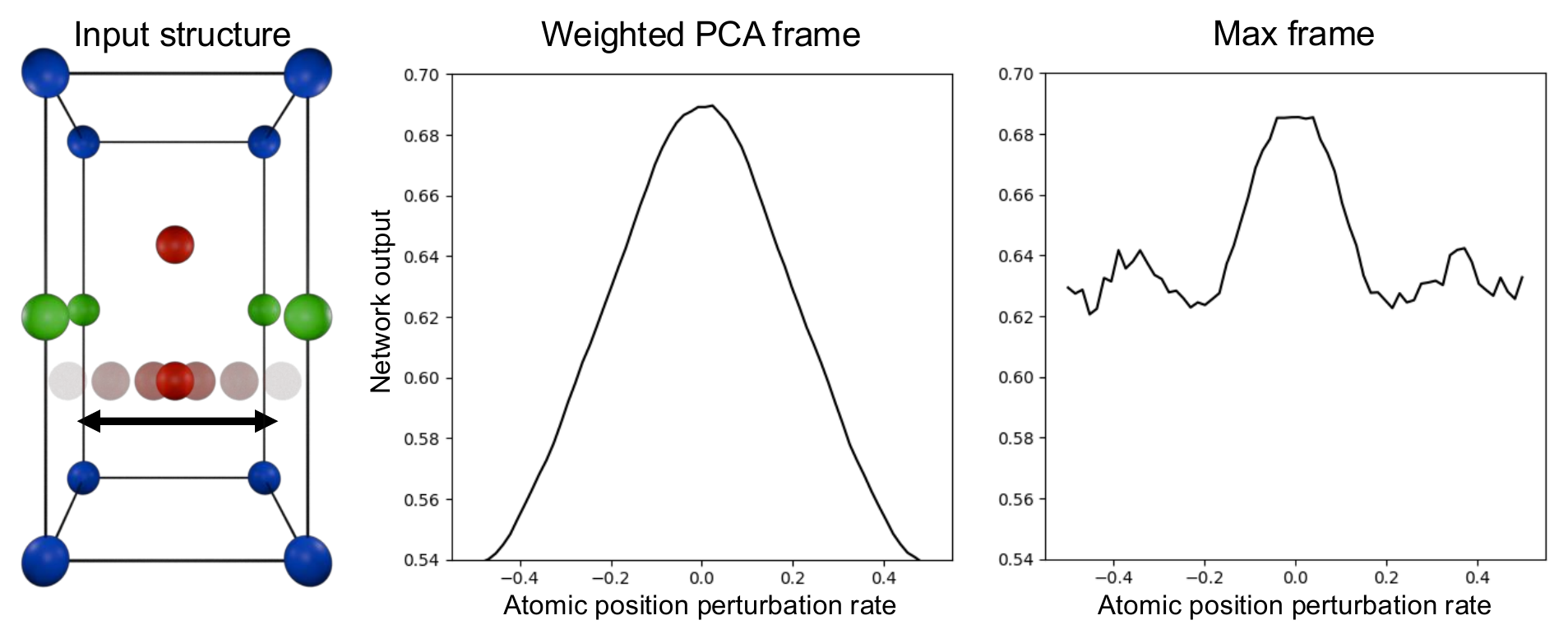}
\caption{\textbf{Continuity under perturbations.} We examine the transitions in the outputs of trained models for a test material under perturbations. Specifically, we use $\text{Be}_2\text{In}\text{Pb}$ (JVASP-70556) from the JARVIS formation energy prediction task and perturb one of the beryllium (Be) atoms along the direction of a lattice vector. The weighted PCA frame model shows a smoother transition compared to the max frame model.
}
\label{fig:appendix_perturbation}
\end{figure}

The results in Figure~\ref{fig:appendix_perturbation} show that the weighted PCA frame model exhibits a significantly smoother transition compared to the max frame model. However, as shown in Tables~\ref{table:results_jarvis} and \ref{table:results_mp}, the weighted PCA frame method has lower performance, indicating that higher continuity does not necessarily translate to better performance. The discontinuous behavior of max frames may have facilitated the early stabilization of frames during training, as discussed in Appendix~\ref{appendix:evolution}, leading to the superior performance.

Meanwhile, the discontinuity of the max frame model becomes more significant with larger perturbations. This trend suggests that the model may have limited generalization to out-of-domain data. The technique of weighted frames proposed by \cite{dym2024continuous} could be applied to improve the continuity of our max frame models.

\end{document}